\documentclass{article}

\PassOptionsToPackage{numbers, compress}{natbib}

\usepackage[final]{unireps_2025}

\usepackage[utf8]{inputenc} 
\usepackage[T1]{fontenc}    
\usepackage{hyperref}       
\usepackage{url}            
\usepackage{booktabs}       
\usepackage{amsfonts}       
\usepackage{nicefrac}       
\usepackage{microtype}      
\usepackage{xcolor}         
\usepackage{float}
\usepackage{graphicx}
\usepackage{amsmath}
\usepackage{tabularx}
\usepackage{placeins}
\usepackage{gensymb}
\usepackage{multirow}
\usepackage{todonotes}
\usepackage{yhmath}

\title{Evaluating Foundation Models' 3D Understanding Through Multi-View Correspondence Analysis}

\author{
  Valentina Lilova\thanks{Equal contribution.} \\
  University of Amsterdam\\
  Amsterdam, 1098 XH \\
  \texttt{valentina.lilova@student.uva.nl} \\
  \And
  Toyesh Chakravorty\footnotemark[1] \\
  University of Amsterdam\\
  Amsterdam, 1098 XH \\
  \texttt{toyesh.chakravorty@student.uva.nl} \\
  \And
  Julian I. Bibo\footnotemark[1] \\
  University of Amsterdam\\
  Amsterdam, 1098 XH \\
  \texttt{julian.bibo@student.uva.nl} \\
  \And
  Emma Boccaletti\footnotemark[1] \\
  University of Amsterdam\\
  Amsterdam, 1098 XH \\
  \texttt{emma.boccaletti@student.uva.nl} \\
  \And
  Brandon Li\footnotemark[1] \\
  University of Amsterdam\\
  Amsterdam, 1098 XH \\
  \texttt{brandon.li@student.uva.nl} \\
  \And
  Lívia Baxová\footnotemark[1] \\
  University of Amsterdam\\
  Amsterdam, 1098 XH \\
  \texttt{livia.baxova@student.uva.nl} \\
  \And
  Cees G. M. Snoek \\
  University of Amsterdam\\
  Amsterdam, 1098 XH \\
  \texttt{c.g.m.snoek@uva.nl} \\
  \And
  Mohammadreza Salehi \\
  University of Amsterdam\\
  Amsterdam, 1098 XH \\
  \texttt{s.salehidehnavi@uva.nl} \\
}

\begin{document}

\maketitle

\begin{abstract}
    
Benchmarking 3D spatial understanding of foundation models is essential for real-world applications such as robotics and autonomous driving. Existing evaluations often rely on downstream fine-tuning with linear heads or task-specific decoders, making it difficult to isolate the intrinsic 3D reasoning ability of pre-trained encoders. In this work, we introduce a novel benchmark for in-context 3D scene understanding that requires no fine-tuning and directly probes the quality of dense visual features. Building on the Hummingbird framework, which evaluates in-context 2D scene understanding, we extend the setup to the 3D Multi-View ImageNet (MVImgNet) dataset. Given a set of images depicting objects at specific camera angles (keys), we benchmark the performance of segmenting novel views (queries) and report the scores in 4 categories of easy, medium, hard, and extreme based on the key-query view contrast. We benchmark 7 state-of-the-art foundation models and show that DINO-based encoders remain competitive across large viewpoint shifts.
Our code is publicly available at \url{https://github.com/ToyeshC/open-hummingbird-3d-eval}.

\end{abstract}

\section{Introduction}
\label{sec:intro}

In recent years, we have seen the rise of foundation models~\cite{opportunities_risks_FoMo, llama, gpt4, qwen3}. These large-scale pre-trained models support many tasks, including visual understanding, which enables direct comparison of their encoders. Despite their impressive capabilities, state-of-the-art Vision Transformers (ViTs) can suffer catastrophic failures when objects are viewed from unusual angles, as demonstrated by recent angle-sensitivity studies~\cite{ruan2023viewpointinvariantvisualrecognitionadversarial}. However, most vision benchmarks emphasize single-view tasks or dense synthesis~\cite{mildenhall2020nerfrepresentingscenesneural, pariza2024near, salehi2023time, salehi2025mosic, ziegler2022self}, leaving segmentation robustness under camera rotations comparatively underexplored.

To address viewpoint variability, recent work has explored in-context learning (ICL) as a means to adapt models to unseen views without retraining. ICL is the ability of a model to perform new tasks by conditioning on a prompt~\cite{explanation_in-context_learning}. An example is Hummingbird~\cite{hummingbird}, a memory-augmented ViT for ICL. Its encoder extracts dense image features and projects them into key-value pairs stored in a dynamic memory. At query time, cross-attention over the keys assigns soft nearest-neighbor (NN) weights to aggregate values and predict novel views in-context. This modular design permits the use of any encoder.

In this work, we evaluate general-purpose ViT encoders \cite{CLIP, DINO, DINOv2, simeoni2025dinov3, SIGLIPv2, radiov2.5, tips_paper} for in-context object segmentation under unseen camera viewpoints. We extend the Hummingbird evaluation framework with the Multi-View ImageNet (MVImgNet) dataset~\cite{Yu_2023_CVPR_mvimgnet}, grouping object views into angular bins. By constructing memory banks consisting of images from specific viewpoints and evaluating segmentation performance on held-out angles, we assess how well different encoders generalize across viewpoint shifts.
Our main contribution combines three elements: (1) pixel-level segmentation, (2) extreme viewpoint changes, and (3) dynamic memory for ICL. Most existing work studies 3D understanding, but there is little evaluation of how consistent the features from frozen 2D foundation models remain when the camera moves to new viewpoints and only a few in-context examples are available.

\section{Related work}
\label{sec:rwork}

Recent developments in view generation have significantly improved the understanding of 3D scene structure in computer vision, setting new benchmarks along the way. 
For example, Neural Radiance Fields (NeRFs)~\cite{mildenhall2020nerfrepresentingscenesneural} achieve state-of-the-art novel view synthesis by optimizing a continuous volumetric function from sparse input images.
While NeRFs addresses generation, other studies investigate recognition in 3D understanding. \citet{ruan2023viewpointinvariantvisualrecognitionadversarial} demonstrate that modern recognition models (ResNets~\cite{residual_learning}, ViTs~\cite{dosovitskiy2020image}, Swin~\cite{swin}, and Masked Autoencoders (MAEs)~\cite{maskedAutoEncoders}) remain sensitive to viewpoint shifts in 3D scenes. With adversarial training, they improve invariance to 3D viewpoint changes beyond standard rotation-based augmentations. In parallel, \citet{shifman2024losttranslationmodernneural} show that state-of-the-art encoders such as CLIP and DINOv2 change predictions under a single-pixel shift in up to 40\% of cases, despite extensive augmentation during training. Although their analysis centers on 2D translation invariance, the findings highlight that robustness to geometric changes, whether in 2D or 3D, remains limited in current vision encoders.

At a higher level, previous benchmarks in 3D computer vision were generally divided into two groups. 
The first group includes those working on traditional 3D recognition tasks, such as item detection (e.g., KITTI, Waymo \cite{electronics14091882, Hwang2023Improvement}) or scene segmentation (e.g., SAI3D, ScanNet \cite{Yin_2024_CVPR}), which typically test explicit 3D architectures or employ classification/bounding box tasks. 
The second focuses on innovative view synthesis approaches, such as NeRF \cite{mildenhall2020nerfrepresentingscenesneural}, which prioritize dense creation rather than robust detection or segmentation under angular deviation.
Beyond these 3D-focused benchmarks, general ICL benchmarks such as the Hummingbird framework \cite{hummingbird} or post-training approaches~\cite{salehi2025mosic, salehi2023time, pariza2024near, ziegler2022self} assess feature transferability only in a 2D setting, and therefore fail to evaluate models under non-linear, wide-angle 3D viewpoint shifts.

Beyond model architectures, multi-view datasets such as MVImgNet and PASCAL3D+ \cite{Pascal3d} offer rich multi-view annotations that support the evaluation of cross-view generalization in object recognition.
The rise of self-supervised Transformers has also opened new possibilities for downstream tasks such as segmentation. In particular, DINOv2~\cite{DINO} produces robust embeddings that transfer effectively to prediction tasks, despite training without labels. More recently, \citet{simeoni2025dinov3} introduced DINOv3, which significantly scales the DINOv2 framework. A critical technical contribution of DINOv3 is the introduction of Gram Anchoring, a mechanism designed to prevent the degradation of dense feature maps during long-duration self-supervised training. By regularizing patch-level similarity structures against a teacher model, this model maintains high-quality spatial features that are crucial for dense prediction tasks.

Nevertheless, a gap remains across these benchmarks and models. Most prior work emphasizes detection, classification, or view-angle estimation, but not the measurement of 3D contextual understanding through multi-view segmentation. 
To address this, our benchmark primarily evaluates general-purpose foundation encoders.
To our knowledge, no existing benchmark systematically evaluates a model’s ability to generalize segmentation across viewpoint shifts using dynamic memory. Our proposed evaluation method aims to address this gap in existing research.

\section{Methodology}
\label{sec:method}
 
We evaluate the view generalization ability of frozen ViT models in semantic segmentation using a non-parametric, retrieval-based framework. 
Our approach builds on the Hummingbird framework~\cite{hummingbird}, which applies ICL to vision tasks and supports the evaluation of spatial perception and semantic understanding. However, Hummingbird does not analyze performance variations across viewpoints changes.
To address this, we introduce a viewpoint binning protocol and a multi-view dataset subset for cross-view robustness analysis. License information for the models, framework, and dataset can be found in Appendix~\ref{app:licenses}.

\subsection{Models}

We evaluate 7 pre-trained ViTs: CLIP~\cite{CLIP}, SigLIP2~\cite{SIGLIPv2}, DINO~\cite{DINO}, DINOv2~\cite{DINOv2}, DINOv3~\cite{simeoni2025dinov3}, C-RADIOv2~\cite{am-radio, radiov2.5}, and TIPS~\cite{tips_paper}.
Most models use a ViT-B/16 backbone, except DINOv2 and TIPS, which have a ViT-B/14 backbone.

\subsection{Inference pipeline}
\label{subsec:method:pipeline}

We follow the Hummingbird~\cite{hummingbird} inference pipeline, where patch-level features from a frozen ViT are stored in a memory bank with one-hot semantic labels. 
During inference, query features are matched to the memory using FAISS\footnote{Facebook AI Similarity Search (FAISS) is an open-source library designed for efficient similarity search and clustering of dense vectors, optimized for high-dimensional feature retrieval \cite{johnson2019billion}.} with cosine similarity and \(k=30\) nearest neighbors. The memory size is passed as an input parameter. Hummingbird's cross-attention decoder aggregates the retrieved labels to produce segmentation predictions.

\subsection{Dataset} 
\label{subec:method:dataset}

Our study is based on MVImgNet, a large-scale dataset with over 6.5 million frames across 238 object categories~\cite{Yu_2023_CVPR_mvimgnet}. Each frame includes a segmentation mask, camera extrinsics, and a reconstructed 3D point cloud. 
This dataset is chosen for three reasons: (1) it contains various annotations across wide viewpoint ranges, enabling view generalization analysis; (2) it includes camera extrinsics that allow precise angular binning; and (3) it has scale and diversity, capturing a broad range of object appearances and spatial configurations.

\subsubsection{Viewpoint binning}
\label{method:dataset:binning}
To study viewpoint robustness, we discretize relative camera angles into 7 bins spanning 0\(^\circ\)--90\(^\circ\), in steps of 15\(^\circ\). 
To aid in this task, we use COLMAP~\cite{COLMAPsfm, COLMAPvote, COLMAPmvs}: a Structure-from-Motion (SfM) and Multi-View Stereo (MVS) pipeline that estimates camera poses and 3D scene geometry from a set of images.
Using COLMAP extrinsics, we compute the relative rotation $R_{rel}$ between each frame and the first frame of the object instance.
For each instance, we select 1 representative frame per bin by choosing the frame with the smallest angular error relative to the bin center. Images and masks are stored per bin for downstream use.
The angular deviation $\theta$ is computed as

\begin{equation}
    \theta = \arccos\left(\frac{\mathrm{trace}(R_{rel}) - 1}{2}\right).
\end{equation}

\subsubsection{Subset construction} 
We curated a subset of MVImgNet, with the goal of selecting categories (classes) with sufficient angular coverage and manageable size for repeated memory-based evaluation. The two criteria used for selecting a category are: (1) the total zipped folder size must be 1--6 GB to ensure feasible data loading and storage; and (2) at least one object instance must span all 7 angular bins with a maximum variance of 6\(^\circ\). Categories that did not meet these criteria were excluded. For example, category 23 (\textit{laptop}) was excluded, as it did not have an instance that reached an angle of 90\(^\circ\). This yielded 15 segmentation classes, excluding the background.\footnote{The 15 categories are ordered as follows: stove, sofa, microwave, bed, toy cat, toy cow, toy dragon, coat rack, guitar stand, ceiling lamp, toilet, sink, strings, broccoli, and durian. More details can be found in Appendix~\ref{App:MVImgNet}.}

\begin{figure}
    \centering
    \includegraphics[width=0.9\linewidth]{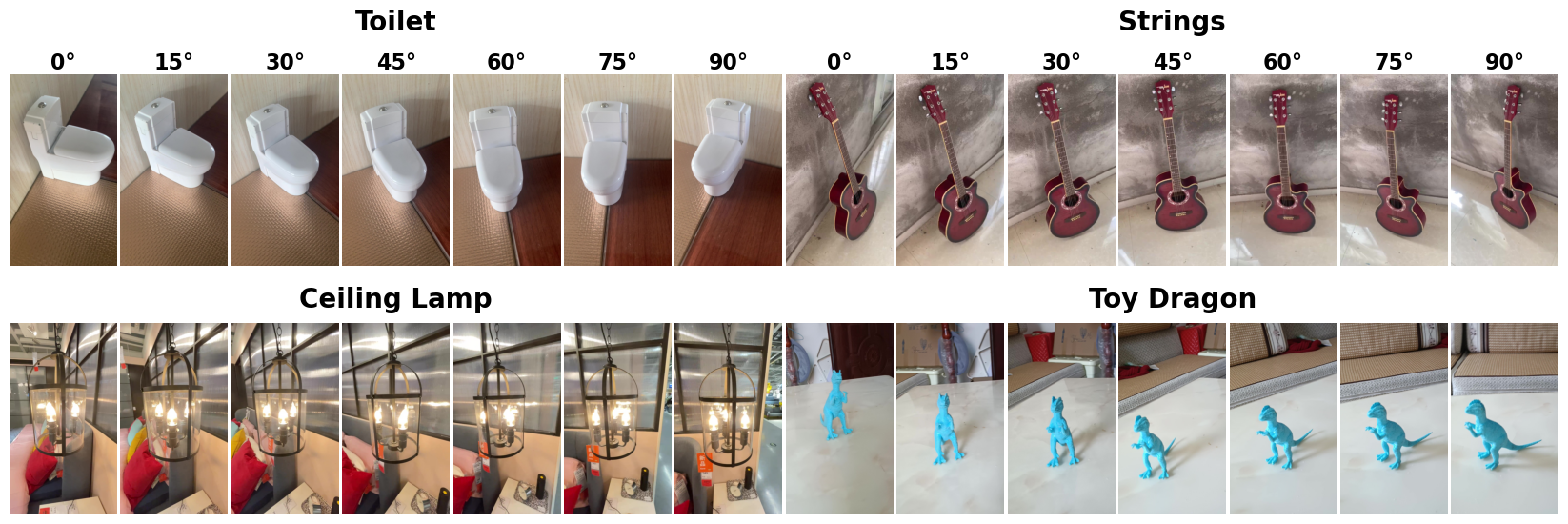}
    \caption{\textbf{Multi-view instances}. Each MVImgNet category contains instances spanning all discretized viewpoint bins between 0\(^\circ\) and 90\(^\circ\). The figure shows example visualizations of 4 instances from the 15 selected classes.}
    \label{fig:small_angles_objects}
\end{figure}

For each category instance, we parsed the COLMAP camera pose data to extract camera extrinsics and compute relative angles.
For each valid instance, we selected the closest frame to each angular bin center and organized the resulting RGB images and masks in a structured directory grouped by category and angle. Additionally, we computed angular selection errors per bin (see~\autoref{tab:angle-selection-stats}).
Examples of different views of several class instances from the resulting dataset are shown in~\autoref{fig:small_angles_objects}. 

\subsection{Model input}

To ensure a fair comparison, we use the largest native input size across all models. This is \(504 \times 504\) for models that use a patch size of 14 and \(512 \times 512\) for the ones using a patch size of 16.
When evaluating models at resolutions higher than those used during pre-training, we resize the pre-trained absolute positional embeddings to match the new patch grid using bicubic interpolation.
All evaluations are performed with a batch size of 4.
The full configuration details are provided in~\autoref{App:Model_Specifics}.

\subsection{Metrics and multi-view evaluation}

The segmentation quality is measured by mean Intersection over Union (mIoU) across all semantic classes.
Each predicted segmentation mask consists of the background class and one of the 15 object classes.
We report 3 metrics for each evaluation setting: the per-class Intersection over Union (IoU), also known as the Jaccard index, mIoU over all 16 classes, and standard deviation. Results are presented per model, reference bin, and validation bin to analyze viewpoint robustness. Although all models perform well on the background class (see~\autoref{fig:expa_bg}), we include it in the mIoU computation for completeness.

\section{Experimental results}

To validate our setup, we first reproduced the Hummingbird evaluation on PASCAL VOC~\cite{Everingham2010Jun} using the official open-source implementation.\footnote{The evaluation repository can be found at \url{https://github.com/vpariza/open-hummingbird-eval}.} The results, shown in~\autoref{tab:hbird-reproduction} of Appendix~\ref{app:reproduction}, match the methodology and configuration of the original paper~\cite{hummingbird}. This confirms that our pipeline is consistent with the benchmark and provides a reliable basis for the experiments that follow.

With the setup validated, we proceed to evaluate viewpoint generalization and robustness across 3 experiments: (A) cross-viewpoint generalization, (B) breaking point analysis, and (C) memory size robustness.

\begin{table}[t]
\footnotesize
\centering
\setlength{\tabcolsep}{3pt}
\caption{\textbf{Difficulties and bin splits}. Difficulty levels are defined by the choice of reference bins (stored in the memory bank) and validation bins (unseen views for testing generalization). An increasing difficulty corresponds to fewer reference bins and larger unseen angular gaps, which makes interpolation and extrapolation progressively harder.}
\begin{tabular}{lcc}
    \toprule
    \textbf{Difficulty} & \textbf{Reference bins} & \textbf{Validation bins} \\
    \midrule
    Easy    & \(0^\circ, 30^\circ, 60^\circ, 90^\circ\) & \(15^\circ, 45^\circ, 75^\circ\) \\
    Medium  & \(0^\circ, 45^\circ, 90^\circ\)           & \(15^\circ, 30^\circ, 60^\circ, 75^\circ\) \\
    Hard    & \(0^\circ, 90^\circ\)                      & \(15^\circ, 30^\circ, 45^\circ, 60^\circ, 75^\circ\) \\
    Extreme & \(0^\circ\)                               & \(15^\circ, 30^\circ, 45^\circ, 60^\circ, 75^\circ, 90^\circ\) \\
    \bottomrule
\end{tabular}
\label{tab:difficulties}
\end{table}

\subsection{Experiment A: cross-viewpoint generalization}
\label{sec:exp_a}

We compare the models in terms of their ability to generalize across viewpoints. The goal is to assess how different pre-training strategies affect performance when only limited reference views are available in the memory.
Specifically, we evaluate how well the models generalize to unseen viewpoints when trained on selected angular bins from the MVImgNet dataset. The reference and validation splits vary across 4 difficulty levels, which are summarized in~\autoref{tab:difficulties}. Each model is evaluated using the mIoU across all 16 classes (background and 15 object categories) and all validation images. 
This setup enables a controlled comparison of each model's ability to interpolate between observed viewpoints and extrapolate to unseen ones.

\subsubsection{Results} 
\label{subsec:exp_a:results}

As shown in~\autoref{tab:exp_a}, DINOv3 outperformed all models across all difficulties. DINO had the second-best performance for the Easy and Medium difficulty levels.
Quantitative results illustrating these trends are shown in~\autoref{fig:expa_all_classes}, with per-class results in Appendix~\ref{app:expa}. 
Across all classes, DINOv3 and DINOv2 maintain stronger segmentation performance as difficulty increases, with DINO remaining competitive under most settings.
CLIP performs close to the DINO models and surpasses DINO under the Extreme setting, while C-RADIOv2, TIPS, and SigLIP2 perform noticeably worse, particularly when reference views are limited.

\begin{table}
\footnotesize
\centering
\caption{\textbf{mIoU scores across difficulty levels}. For each model, we report the average mIoU and standard deviation over the 4 difficulty levels. The memory size used is 1,024k.
DINOv3 achieves the best performance across all difficulties, followed by DINOv2 and DINO, of which the latter outperformed DINOv2 in the Easy and Medium setups (where multiple reference views are available). DINOv2 significantly outperformed DINO under the Extreme case with only a single reference bin. 
}
\begin{tabular}{lcccc}
    \toprule
    \textbf{Model} & \textbf{Easy} & \textbf{Medium} & \textbf{Hard} & \textbf{Extreme} \\
    \midrule
    CLIP ViT-B/16 & 0.755 $\pm$ 0.130 & 0.748 $\pm$ 0.134 & 0.734 $\pm$ 0.140 & 0.701 $\pm$ 0.149 \\
    DINO ViT-B/16 & 0.782 $\pm$ 0.132 & 0.774 $\pm$ 0.137 & 0.748 $\pm$ 0.153 & 0.686 $\pm$ 0.171 \\
    DINOv2 ViT-B/14 & 0.763 $\pm$ 0.136 & 0.758 $\pm$ 0.139 & 0.748 $\pm$ 0.143 & 0.728 $\pm$ 0.154 \\
    DINOv3 ViT-B/16 & \textbf{0.809 $\pm$ 0.130} & \textbf{0.803 $\pm$ 0.133} & \textbf{0.790 $\pm$ 0.144} & \textbf{0.773 $\pm$ 0.158} \\
    C-RADIOv2 ViT-B/16-CPE & 0.653 $\pm$ 0.129 & 0.636 $\pm$ 0.132 & 0.592 $\pm$ 0.141 & 0.506 $\pm$ 0.150 \\
    SigLIP2 B/16-512 & 0.564 $\pm$ 0.150 & 0.551 $\pm$ 0.153 & 0.530 $\pm$ 0.157 & 0.481 $\pm$ 0.152 \\
    TIPS ViT-B/14-HR & 0.667 $\pm$ 0.137 & 0.647 $\pm$ 0.146 & 0.588 $\pm$ 0.162 & 0.462 $\pm$ 0.169 \\
    \bottomrule
\end{tabular}
\label{tab:exp_a}
\end{table}

\subsubsection{Discussion} 
\label{subsec:exp_a:discussion}

Although none of the models were explicitly trained for viewpoint understanding, the results show clear differences in their ability to generalize across views. As seen in the previous section, DINOv3 performs best, followed by DINOv2 and DINO.
This suggests that the features of the DINO models, compared to other models, are more stable under viewpoint changes and better support shape-related cues.
This consistency may arise from their self-supervised training objective, which explicitly encourages feature alignment across multiple augmentations of the same image and more stable relationships between local and global representations.
DINOv3 further benefits from large-scale pre-training and the Gram Anchoring mechanism~\cite{simeoni2025dinov3}, which might contribute to more stable local features and stronger geometric consistency under viewpoint changes.

While DINO performed better under the easier difficulties, where multiple views of the object were available, DINOv2 obtained a higher mIoU under the Extreme difficulty with only 1 reference bin. 
One possible interpretation is that DINO's simpler self-distillation loss favors interpolation across views, whereas DINOv2's larger-scale self-supervised pre-training 
better supports generalization to more extreme viewpoints.
CLIP generally ranked third, reflecting some ability to encode visual-semantic regularities despite being trained for image-text alignment rather than geometric consistency.  
By comparison, SigLIP2, TIPS, and C-RADIOv2 performed substantially worse, suggesting that objectives centered on semantic matching or mixed supervision may weaken part-level consistency needed for viewpoint-robust segmentation.

\begin{figure}
  \centering 
  \includegraphics[width=\linewidth]{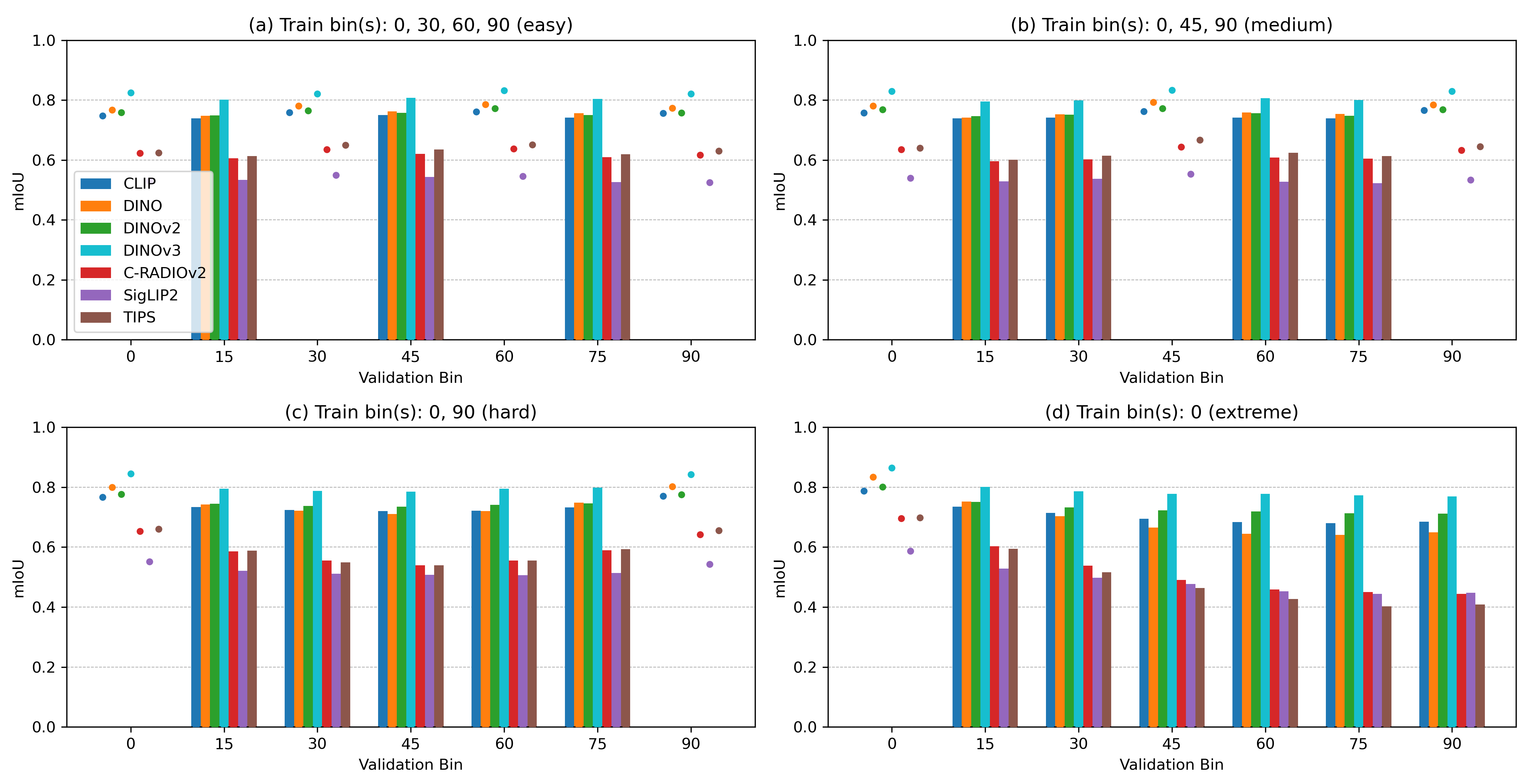}
  \caption{\textbf{Segmentation performance across viewpoint bins}. Each subplot (a)--(d) corresponds to a different difficulty level. Bars show mIoU scores on unseen validation bins, while dots show performance on the reference bins stored in the memory bank. The figure illustrates how segmentation accuracy declines as the validation angles move farther from the reference viewpoints. DINOv3, DINOv2, DINO, and CLIP maintain higher, more stable performance as angular distance increases, whereas other models degrade more quickly. 
  Overall, the trend highlights how self-supervised pre-training leads to smoother cross-view generalization.}

  \label{fig:expa_all_classes}
\end{figure}

\subsection{Experiment B: breaking point analysis}
\label{sec:exp_b}

In this experiment, we analyze whether the models experience a sudden failure in viewpoint generalization, and if so, at what angle. The goal is to identify the angular range within which performance remains stable under the Extreme difficulty. In particular, we perform a breaking point analysis under the Extreme setting from experiment~A, where only a single reference bin (0$^\circ$) is available in the memory and models are evaluated across the remaining bins. We define the breaking point as the earliest validation bin where a model's normalized mIoU score $\tilde{m}_i$ drops significantly compared to the previous bin ($\tilde{m}_{i-1}$). Specifically, we compute the performance drop $\Delta_i$ per bin $i$ as

\begin{align}
    \Delta_i &= \tilde{m}_i - \tilde m_{i-1}, \\
    \label{eq:norm_miou}
    \tilde m_i &= \frac{m_i}{m_{0^\circ}},
\end{align}

where the normalization by the 0$^\circ$ bin (\autoref{eq:norm_miou}) ensures that performance drops reflect viewpoint sensitivity rather than overall model scale or performance.

A breaking point is recorded at bin $i$ if $\Delta_i \leq -0.1$, indicating a relative drop of 10\% or more. This analysis highlights each model's resistance to viewpoint shifts and identifies the angular range within which performance remains stable. We also report the normalized mIoU degradation curves as the viewpoint angle increases. 

\subsubsection{Results} 
\label{subsec:exp_b:results}

\begin{figure}
    \centering
    \includegraphics[width=0.6\linewidth]{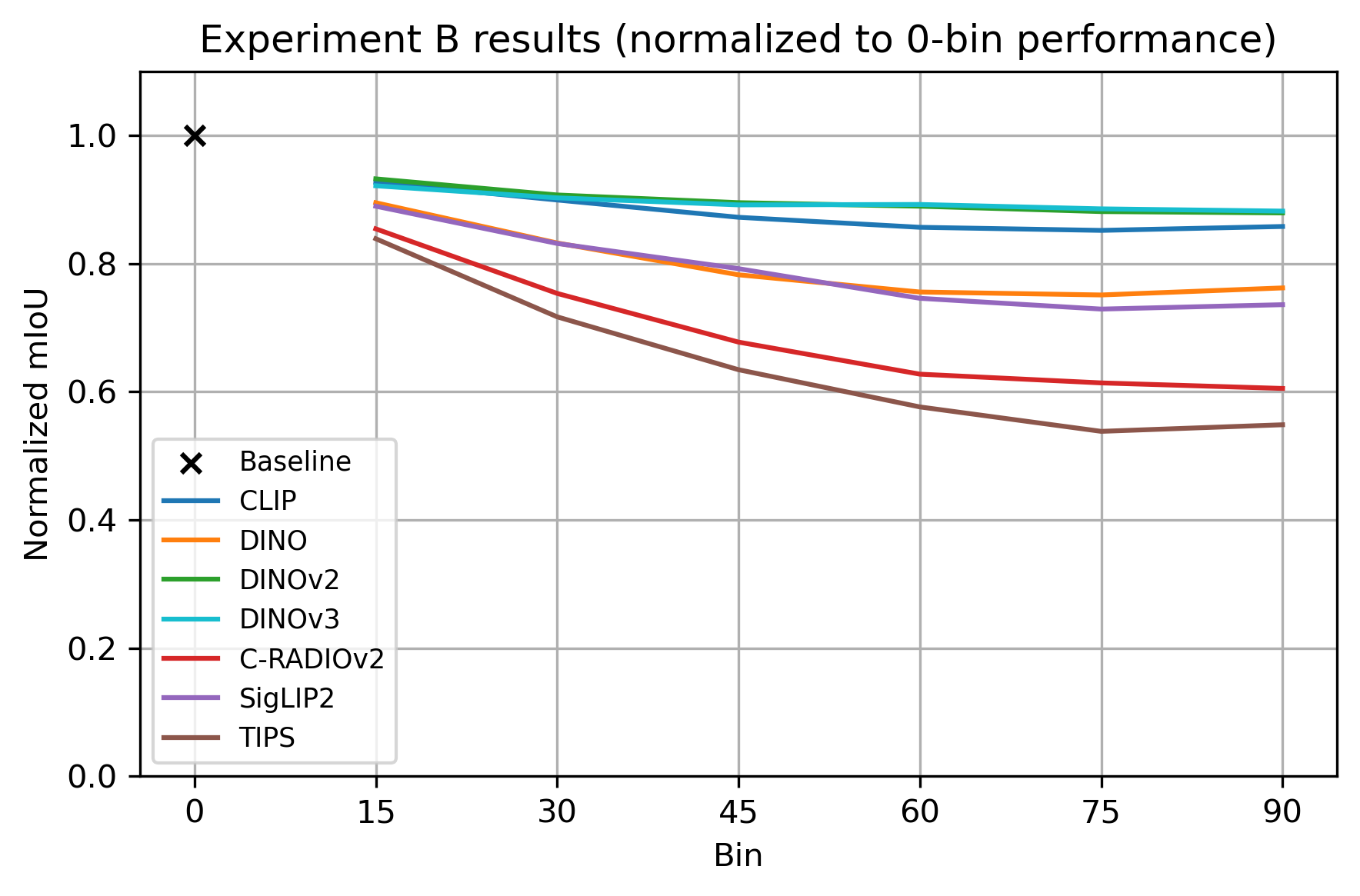}
    \caption{\textbf{Normalized mIoU under viewpoint shifts}. 
    Each curve shows the normalized mIoU of a model relative to its performance in the $0^\circ$ reference bin. The horizontal axis corresponds to increasing viewpoint angles (validation bins). All models exhibit a monotonic decrease with diminishing slope, indicating that the steepest performance degradation occurs at the initial viewpoint shifts, followed by progressively flatter declines at larger angles. DINOv3, DINOv2, and CLIP degrade smoothly, followed by DINO and SigLIP2, whereas TIPS and C-RADIOv2 display sharper early drops, suggesting weaker robustness to small viewpoint changes.
    }
    \label{fig:expb_norm}
\end{figure}

As shown in \autoref{fig:expb_norm}, all models exhibit a monotonic decrease in normalized mIoU as the validation viewpoint angle increases. The drop is steepest at small viewpoint changes and gradually flattens at larger angles, indicating that most performance degradation occurs at the initial viewpoint shifts.

TIPS is the only model that reaches a breaking point, occurring at the $30^\circ$ bin (see~\autoref{tab:expb_breakpoints}), with a normalized mIoU drop of $-0.1148$. C-RADIOv2 shows a relatively sharp early decline but remains below the breaking-point threshold. DINOv3, DINOv2, and CLIP exhibit the smoothest degradation, followed by DINO and SigLIP2.

\begin{table}
\setlength{\tabcolsep}{4pt}
\footnotesize
\centering
\caption{\textbf{Breaking points}. For each model, we report the breaking point validation bin where the normalized mIoU drop exceeds 10\% (\(\Delta_i \leq -0.1\)) relative to the previous bin. ``None'' indicates gradual degradation. Only TIPS reaches a breaking point. C-RADIOv2 is close to having a breaking point, but it doesn't pass the threshold as per our definition. DINO, DINOv2, CLIP, and SigLIP2 maintain smoother decline curves. The biggest drops for all models occur between 15\(^\circ\) and 30\(^\circ\) as seen in \autoref{fig:expb_norm}.
}

\begin{tabular}{lccc}
\toprule
\textbf{Model} & \textbf{Breaking point bin} & \textbf{Biggest drop} \\
\midrule
CLIP ViT-B/16 & None & \(-0.0267\) \\
DINO ViT-B/16 & None & \(-0.0559\) \\
DINOv2 ViT-B/14 & None & \(-0.0273\) \\
DINOv3 ViT-B/16 & None & \(-0.0214\) \\
C-RADIOv2 ViT-B/16-CPE & None & \(-0.0969\) \\
SigLIP2 B/16-512 & None & \(-0.0550\) \\
TIPS ViT-B/14-HR & 30 & \(-0.1148\) \\
\bottomrule
\end{tabular}

\label{tab:expb_breakpoints}
\end{table}

\subsubsection{Discussion} 
\label{subsec:exp_b:discussion}


The breaking point analysis highlights differences in robustness across models and shows how their pre-training objectives affect stability under viewpoint shifts. DINOv2 and DINOv3 are the most stable, maintaining gradual degradation without sudden drops. CLIP shows a similar trend, with slightly weaker overall performance. 
The normalized performances of DINO and SigLIP2 also decrease steadily, with a slight increase at the 90$^\circ$ bin. 
Interestingly, CLIP and SigLIP2 preserve some resilience despite being trained for image-text alignment rather than spatial structure. Although C-RADIOv2 does not reach a breaking point, it comes very close to one, which we could attribute to its mixed supervision weakening the reliability of part-level representations.  
In contrast, TIPS exhibits a clear breakdown, suggesting that its learned features are less robust when faced with unseen viewpoints.
This instability may stem from its focus on semantic alignment rather than spatial consistency. 
Relative to the baseline that includes validation bins in the memory, the observed drops can be attributed to missing reference views rather than to instability in the evaluation pipeline.
Overall, the results suggest that self-supervised encoders such as the DINO-based models produce more geometry-aware features under strong viewpoint shifts, while multimodal or distilled approaches may be more brittle. DINOv2 and DINOv3's stability make them the most reliable choice for applications that must handle abrupt perspective changes.

\subsection{Experiment C: memory size robustness}
\label{sec:exp_c}

In this final experiment, we analyze how memory bank size affects viewpoint generalization, comparing the performance gains across difficulty levels against the added computational cost.
We follow the Hummingbird paper~\cite{hummingbird} and vary the number of entries in the memory bank. Three memory sizes are evaluated: 320k, 640k, and 1,024k. For each size, we report the segmentation accuracy.

\subsubsection{Results}
\label{subsec:exp_c:results}

The results in~\autoref{tab:memory_impact_performance} show that increasing the memory bank generally improves mIoU, with the largest gains observed under the Easy difficulty. Doubling the memory from 320k to 640k increases performance by approximately 0.030 mIoU on Easy, while a further increase to 1,024k yields a smaller improvement of 0.017 (see~\autoref{tab:hbird_memory_gain_combined} in Appendix~\ref{app.ExpC}). 

DINOv3 achieves the highest absolute performance across all memory sizes, followed by DINO and DINOv2. Weaker models such as C-RADIOv2, SigLIP2, and TIPS benefit more 
from additional memory, improving by more than 0.048 mIoU when increasing the memory from 320k to 1,024k, with most of the gain coming from the first doubling of memory.

Increasing the memory bank introduces additional practical considerations, as documented in Appendix~\ref{app.ExpC}.


\begin{table}
\centering
\footnotesize
\caption{\textbf{Comparison of memory-bank size on segmentation performance}. Mean mIoU scores and standard deviations are shown for 3 memory sizes ((a)~320k, (b)~640k, (c)~1,024k) and 4 difficulty levels. Bold values mark the best score per difficulty-memory configuration. Increasing memory generally improves performance, especially for weaker encoders such as SigLIP2. The DINO models remain the top performers across all settings, but have minimal gains past 640k. 
}

\begin{tabular}{lcccc}
    \toprule
    \textbf{Model} & \textbf{Easy} & \textbf{Medium} & \textbf{Hard} & \textbf{Extreme} \\
    \midrule
    \multicolumn{5}{c}{(a) Memory: 320k} \\
    \midrule
    CLIP ViT-B/16 & 0.729 $\pm$ 0.138 & 0.725 $\pm$ 0.141 & 0.710 $\pm$ 0.149 & 0.681 $\pm$ 0.156 \\
    DINO ViT-B/16 & 0.741 $\pm$ 0.147 & 0.736 $\pm$ 0.150 & 0.712 $\pm$ 0.163 & 0.656 $\pm$ 0.178 \\
    DINOv2 ViT-B/14 & 0.738 $\pm$ 0.145 & 0.736 $\pm$ 0.146 & 0.726 $\pm$ 0.152 & 0.709 $\pm$ 0.158 \\
    DINOv3 ViT-B/16 & \textbf{0.793 $\pm$ 0.139} & \textbf{0.788 $\pm$ 0.141} & \textbf{0.780 $\pm$ 0.149} & \textbf{0.768 $\pm$ 0.157} \\
    C-RADIOv2 ViT-B/16-CPE & 0.588 $\pm$ 0.136 & 0.579 $\pm$ 0.139 & 0.539 $\pm$ 0.146 & 0.467 $\pm$ 0.149 \\
    SigLIP2 B/16-512 & 0.506 $\pm$ 0.154 & 0.501 $\pm$ 0.156 & 0.483 $\pm$ 0.157 & 0.443 $\pm$ 0.149 \\
    TIPS ViT-B/14-HR & 0.600 $\pm$ 0.161 & 0.590 $\pm$ 0.166 & 0.539 $\pm$ 0.178 & 0.437 $\pm$ 0.175 \\

    \midrule
    \multicolumn{5}{c}{(b) Memory: 640k} \\
    \midrule
    CLIP ViT-B/16 & 0.745 $\pm$ 0.132 & 0.740 $\pm$ 0.136 & 0.727 $\pm$ 0.141 & 0.694 $\pm$ 0.150 \\
    DINO ViT-B/16 & 0.768 $\pm$ 0.137 & 0.761 $\pm$ 0.141 & 0.736 $\pm$ 0.156 & 0.676 $\pm$ 0.173 \\
    DINOv2 ViT-B/14 & 0.754 $\pm$ 0.138 & 0.750 $\pm$ 0.142 & 0.740 $\pm$ 0.147 & 0.721 $\pm$ 0.155 \\   
    DINOv3 ViT-B/16 & \textbf{0.803 $\pm$ 0.133} & \textbf{0.798 $\pm$ 0.136} & \textbf{0.787 $\pm$ 0.145} & \textbf{0.771 $\pm$ 0.157} \\
    C-RADIOv2 ViT-B/16-CPE & 0.629 $\pm$ 0.132 & 0.615 $\pm$ 0.135 & 0.572 $\pm$ 0.142 & 0.491 $\pm$ 0.150 \\
    SigLIP2 B/16-512 & 0.541 $\pm$ 0.152 & 0.531 $\pm$ 0.154 & 0.511 $\pm$ 0.157 & 0.466 $\pm$ 0.149 \\
    TIPS ViT-B/14-HR & 0.644 $\pm$ 0.144 & 0.628 $\pm$ 0.153 & 0.571 $\pm$ 0.167 & 0.453 $\pm$ 0.170 \\

    \midrule
    \multicolumn{5}{c}{(c) Memory: 1,024k} \\
    \midrule
    CLIP ViT-B/16 & 0.755 $\pm$ 0.130 & 0.748 $\pm$ 0.134 & 0.734 $\pm$ 0.140 & 0.701 $\pm$ 0.149 \\
    DINO ViT-B/16 & 0.782 $\pm$ 0.132 & 0.774 $\pm$ 0.137 & 0.748 $\pm$ 0.153 & 0.686 $\pm$ 0.171 \\
    DINOv2 ViT-B/14 & 0.763 $\pm$ 0.136 & 0.758 $\pm$ 0.139 & 0.748 $\pm$ 0.143 & 0.728 $\pm$ 0.154 \\
    DINOv3 ViT-B/16 & \textbf{0.809 $\pm$ 0.130} & \textbf{0.803 $\pm$ 0.133} & \textbf{0.790 $\pm$ 0.144} & \textbf{0.773 $\pm$ 0.158} \\
    C-RADIOv2 ViT-B/16-CPE & 0.653 $\pm$ 0.129 & 0.636 $\pm$ 0.132 & 0.592 $\pm$ 0.141 & 0.506 $\pm$ 0.150 \\
    SigLIP2 B/16-512 & 0.564 $\pm$ 0.150 & 0.551 $\pm$ 0.153 & 0.530 $\pm$ 0.157 & 0.481 $\pm$ 0.152 \\
    TIPS ViT-B/14-HR & 0.667 $\pm$ 0.137 & 0.647 $\pm$ 0.146 & 0.588 $\pm$ 0.162 & 0.462 $\pm$ 0.169 \\
    
    \bottomrule
\end{tabular}  
\label{tab:memory_impact_performance}
\end{table}

\subsubsection{Discussion} 
\label{subsec:exp_c:discussion}

These results confirm that larger memory banks improve generalization, but the benefits vary across models. Less robust encoders such as C-RADIOv2, SigLIP2, and TIPS gain the most from additional memory, while DINOv3, DINO, and DINOv2 already perform strongly even with smaller banks. The diminishing gains beyond 640k and the higher computational cost suggest that the intermediate configuration offers the best trade-off. This shows that memory scaling can compensate for weaker representations, but stable self-supervised features remain more effective overall.
\section{Qualitative analysis of segmentations}
\label{sec:qualitative_analysis}

In addition to quantitative evaluations, we conducted a qualitative inspection of predicted segmentation masks to better understand model behavior under the large-angle changes of the Extreme difficulty. The illustrations mentioned in this section can be found in Appendix~\ref{app:quali_analysis}. 

\subsection{Localized shape recovery}
As illustrated in \autoref{fig:gt_pred_input_dino_triplet}, models generally segment the global shape of an object even when the viewpoint differs substantially from any reference view. Although predictions (center) are not perfectly pixel-aligned with the ground truth (left), the overall contour and structure are typically preserved. Minor over-extensions at the ends of the object (e.g., protruding elements like a tail) are common, reflecting the limits of fine-grained spatial recall under large angular deviations.

\subsection{Prediction versus ground truth overlays}

Overlays of the predicted mask (center) versus the ground truth (left) on the input image, such as those in \autoref{fig:overlay_gt_pred_dino}, reveal systematic patterns: models tend to underestimate boundaries in regions with strong self-occlusion or low illumination, while occasionally oversegmenting areas with visually similar textures or shadows. Despite these deviations, the predictions largely align with the main visual cues of the target object, suggesting strong local feature retrieval even from distant viewpoints.


\subsection{Prediction versus ground truth}
Color-coded difference maps (e.g., \autoref{fig:overlay_gt_pred_dino} (right), or \autoref{fig:gt_vs_pred_triceratops}) show that most misalignments occur near object borders or in thin, high-curvature regions (e.g., legs, tails). This indicates that memory retrieval may
struggle with fine-grained spatial resolution under Extreme difficulty views.
As illustrated in Figures~\ref{fig:12_bed} and~\ref{fig:bed}--\ref{fig:bed6} for the \textit{bed} class, some categories may have imprecise or inconsistent ground truth masks which contribute to measured errors, sometimes penalizing predictions that are visually more plausible. 
Such cases highlight that annotation quality can limit the reliability of quantitative evaluation, since metrics are tied to
imperfect labels rather than perceptual correctness.


\subsection{Failures}
Failure cases, such as the \textit{toy dragon} shown in \autoref{fig:gt_vs_pred_triceratops}, typically arise when the model retrieves visually similar but geometrically inconsistent patches. This may lead to boundary inflation. Such failures may explain the sharp performance drops observed in certain models (e.g., TIPS) at larger angular deviations, where cross-view consistency becomes more challenging.

\section{Conclusion}
\label{sec:conclusion}

We studied the ability of frozen ViT encoders to generalize across unseen camera viewpoints in an in-context segmentation setting, using the Hummingbird architecture and MVImgNet dataset. Our benchmark covered three aspects: controlled evaluation under view shifts, robustness through breaking point analysis, and the effect of memory bank size. Our findings show that self-supervised encoders, particularly DINO, DINOv2, and DINOv3, achieve higher performance under viewpoint changes than multimodal and mixed-supervision approaches.
Increasing the memory bank size benefits weaker models and offers diminishing returns for stronger encoders.

Taken together, these results highlight the strengths and limitations of current ViTs for multi-view perception. 
While self-supervised training objectives are associated with improved robustness, none of the models are immune to viewpoint-induced degradation. 
Addressing this gap will likely require objectives or architectures 
that more explicitly account for 3D or viewpoint consistency.
Future work could include extending this evaluation to multi-object or multi-class scenes, wider angular ranges, and compound rotations, providing a broader testbed for building truly viewpoint-robust representations.

\bibliographystyle{plainnat}
\bibliography{references}

@inproceedings{dosovitskiy2020image,
    title={{An Image is Worth 16x16 Words: Transformers for Image Recognition at Scale}},
    author={Dosovitskiy, Alexey and Beyer, Lucas and Kolesnikov, Alexander and Weissenborn, Dirk and Zhai, Xiaohua and Unterthiner, Thomas and Dehghani, Mostafa and Minderer, Matthias and Heigold, Georg and Gelly, Sylvain and Uszkoreit, Jakoba and Houlsby, Neil},
    booktitle={International Conference on Learning Representations},
    year={2020}
}

@InProceedings{Yu_2023_CVPR_mvimgnet,
    author    = {Yu, Xianggang and Xu, Mutian and Zhang, Yidan and Liu, Haolin and Ye, Chongjie and Wu, Yushuang and Yan, Zizheng and Zhu, Chenming and Xiong, Zhangyang and Liang, Tianyou and Chen, Guanying and Cui, Shuguang and Han, Xiaoguang},
    title     = {MVImgNet: A Large-Scale Dataset of Multi-View Images},
    booktitle = {Proceedings of the IEEE/CVF Conference on Computer Vision and Pattern Recognition (CVPR)},
    month     = {June},
    year      = {2023},
    pages     = {9150-9161}
}

@inproceedings{hummingbird,
 author={Bala{\v{z}}evi{\'c}, Ivana and Steiner, David and Parthasarathy, Nikhil and Arandjelovi{\'c}, Relja and H{\'e}naff, Olivier J.},
 booktitle = {Advances in Neural Information Processing Systems},
 pages = {63758--63778},
 publisher = {Curran Associates, Inc.},
 title = {Towards In-context Scene Understanding},
 url = {https://proceedings.neurips.cc/paper_files/paper/2023/file/c94a632545000531f0b47000e9caa5b6-Paper-Conference.pdf},
 volume = {36},
 year = {2023}
}

@InProceedings{tips_paper, 
    Title = {{TIPS: Text-Image Pretraining with Spatial Awareness}}, 
    Author = {Maninis, Kevis-Kokitsi and Chen, Kaifeng and Ghosh, Soham and Karpur, Arjun and Chen, Koert and Xia, Ye and Cao, Bingyi and Salz, Daniel and Han, Guangxing and Dlabal, Jan and Gnanapragasam, Dan and Seyedhosseini, Mojtaba and Zhou, Howard and Araujo, André}, 
    Booktitle = {ICLR}, 
    year = {2025} 
}

@InProceedings{CLIP,
  title = 	 {Learning Transferable Visual Models From Natural Language Supervision},
  author =       {Radford, Alec and Kim, Jong Wook and Hallacy, Chris and Ramesh, Aditya and Goh, Gabriel and Agarwal, Sandhini and Sastry, Girish and Askell, Amanda and Mishkin, Pamela and Clark, Jack and Krueger, Gretchen and Sutskever, Ilya},
  booktitle = 	 {Proceedings of the 38th International Conference on Machine Learning},
  pages = 	 {8748--8763},
  year = 	 {2021},
  volume = 	 {139},
  series = 	 {Proceedings of Machine Learning Research},
  month = 	 {18--24 Jul},
  publisher =    {PMLR},
  url = 	 {https://proceedings.mlr.press/v139/radford21a.html},
}

@INPROCEEDINGS{radiov2.5,
  author={Heinrich, Greg and Ranzinger, Mike and Yin, Hongxu Danny and Lu, Yao and Kautz, Jan and Tao, Andrew and Catanzaro, Bryan and Molchanov, Pavlo},
  booktitle={2025 IEEE/CVF Conference on Computer Vision and Pattern Recognition (CVPR)}, 
  title={RADIOv2.5: Improved Baselines for Agglomerative Vision Foundation Models}, 
  year={2025},
  volume={},
  number={},
  pages={22487-22497},
  doi={10.1109/CVPR52734.2025.02094}}

@InProceedings{am-radio,
    author    = {Ranzinger, Mike and Heinrich, Greg and Kautz, Jan and Molchanov, Pavlo},
    title     = {AM-RADIO: Agglomerative Vision Foundation Model Reduce All Domains Into One},
    booktitle = {Proceedings of the IEEE/CVF Conference on Computer Vision and Pattern Recognition (CVPR)},
    month     = {June},
    year      = {2024},
    pages     = {12490-12500}
}

@INPROCEEDINGS{DINO,
  author={Caron, Mathilde and Touvron, Hugo and Misra, Ishan and Jegou, Hervé and Mairal, Julien and Bojanowski, Piotr and Joulin, Armand},
  booktitle={2021 IEEE/CVF International Conference on Computer Vision (ICCV)}, 
  title={Emerging Properties in Self-Supervised Vision Transformers}, 
  year={2021},
  volume={},
  number={},
  pages={9630-9640},
  doi={10.1109/ICCV48922.2021.00951}}

@misc{DINOv2,
    title={DINOv2: Learning Robust Visual Features without Supervision}, 
    author={Maxime Oquab and Timothée Darcet and Théo Moutakanni and Huy Vo and Marc Szafraniec and Vasil Khalidov and Pierre Fernandez and Daniel Haziza and Francisco Massa and Alaaeldin El-Nouby and Mahmoud Assran and Nicolas Ballas and Wojciech Galuba and Russell Howes and Po-Yao Huang and Shang-Wen Li and Ishan Misra and Michael Rabbat and Vasu Sharma and Gabriel Synnaeve and Hu Xu and Hervé Jegou and Julien Mairal and Patrick Labatut and Armand Joulin and Piotr Bojanowski},
    year={2024},
    eprint={2304.07193},
    archivePrefix={arXiv},
    primaryClass={cs.CV},
    url={https://arxiv.org/abs/2304.07193}, 
}

@inproceedings{COLMAPsfm,
    author={Sch\"{o}nberger, Johannes Lutz and Frahm, Jan-Michael},
    title={Structure-from-Motion Revisited},
    booktitle={Conference on Computer Vision and Pattern Recognition (CVPR)},
    year={2016},
}

@inproceedings{COLMAPmvs,
    author={Sch\"{o}nberger, Johannes Lutz and Zheng, Enliang and Pollefeys, Marc and Frahm, Jan-Michael},
    title={Pixelwise View Selection for Unstructured Multi-View Stereo},
    booktitle={European Conference on Computer Vision (ECCV)},
    year={2016},
}

@inproceedings{COLMAPvote,
    author={Sch\"{o}nberger, Johannes Lutz and Price, True and Sattler, Torsten and Frahm, Jan-Michael and Pollefeys, Marc},
    title={A Vote-and-Verify Strategy for Fast Spatial Verification in Image Retrieval},
    booktitle={Asian Conference on Computer Vision (ACCV)},
    year={2016},
}

@article{mildenhall2020nerfrepresentingscenesneural,
author = {Mildenhall, Ben and Srinivasan, Pratul P. and Tancik, Matthew and Barron, Jonathan T. and Ramamoorthi, Ravi and Ng, Ren},
title = {NeRF: representing scenes as neural radiance fields for view synthesis},
year = {2021},
issue_date = {January 2022},
publisher = {Association for Computing Machinery},
address = {New York, NY, USA},
volume = {65},
number = {1},
issn = {0001-0782},
url = {https://doi.org/10.1145/3503250},
doi = {10.1145/3503250},
journal = {Commun. ACM},
month = dec,
pages = {99–106},
numpages = {8}
}

@InProceedings{ruan2023viewpointinvariantvisualrecognitionadversarial,
    author    = {Ruan, Shouwei and Dong, Yinpeng and Su, Hang and Peng, Jianteng and Chen, Ning and Wei, Xingxing},
    title     = {Towards Viewpoint-Invariant Visual Recognition via Adversarial Training},
    booktitle = {Proceedings of the IEEE/CVF International Conference on Computer Vision (ICCV)},
    month     = {October},
    year      = {2023},
    pages     = {4709-4719}
}

@INPROCEEDINGS{Pascal3d,
    author={Xiang, Yu and Mottaghi, Roozbeh and Savarese, Silvio},
    booktitle={IEEE Winter Conference on Applications of Computer Vision}, 
    title={Beyond PASCAL: A benchmark for 3D object detection in the wild}, 
    year={2014},
    volume={},
    number={},
    pages={75-82},
    doi={10.1109/WACV.2014.6836101}
}

@misc{SIGLIPv2,
    title={SigLIP 2: Multilingual Vision-Language Encoders with Improved Semantic Understanding, Localization, and Dense Features}, 
    author={Michael Tschannen and Alexey Gritsenko and Xiao Wang and Muhammad Ferjad Naeem and Ibrahim Alabdulmohsin and Nikhil Parthasarathy and Talfan Evans and Lucas Beyer and Ye Xia and Basil Mustafa and Olivier Hénaff and Jeremiah Harmsen and Andreas Steiner and Xiaohua Zhai},
    year={2025},
    eprint={2502.14786},
    archivePrefix={arXiv},
    primaryClass={cs.CV},
    url={https://arxiv.org/abs/2502.14786}, 
}

@misc{explanation_in-context_learning,
    title={An Explanation of In-context Learning as Implicit Bayesian Inference}, 
    author={Sang Michael Xie and Aditi Raghunathan and Percy Liang and Tengyu Ma},
    year={2022},
    eprint={2111.02080},
    archivePrefix={arXiv},
    primaryClass={cs.CL},
    url={https://arxiv.org/abs/2111.02080}, 
}

@misc{opportunities_risks_FoMo,
    title={On the Opportunities and Risks of Foundation Models}, 
    author={Rishi Bommasani and Drew A. Hudson and Ehsan Adeli and Russ Altman and Simran Arora and Sydney von Arx and Michael S. Bernstein and Jeannette Bohg and Antoine Bosselut and Emma Brunskill and Erik Brynjolfsson and Shyamal Buch and Dallas Card and Rodrigo Castellon and Niladri Chatterji and Annie Chen and Kathleen Creel and Jared Quincy Davis and Dora Demszky and Chris Donahue and Moussa Doumbouya and Esin Durmus and Stefano Ermon and John Etchemendy and Kawin Ethayarajh and Li Fei-Fei and Chelsea Finn and Trevor Gale and Lauren Gillespie and Karan Goel and Noah Goodman and Shelby Grossman and Neel Guha and Tatsunori Hashimoto and Peter Henderson and John Hewitt and Daniel E. Ho and Jenny Hong and Kyle Hsu and Jing Huang and Thomas Icard and Saahil Jain and Dan Jurafsky and Pratyusha Kalluri and Siddharth Karamcheti and Geoff Keeling and Fereshte Khani and Omar Khattab and Pang Wei Koh and Mark Krass and Ranjay Krishna and Rohith Kuditipudi and Ananya Kumar and Faisal Ladhak and Mina Lee and Tony Lee and Jure Leskovec and Isabelle Levent and Xiang Lisa Li and Xuechen Li and Tengyu Ma and Ali Malik and Christopher D. Manning and Suvir Mirchandani and Eric Mitchell and Zanele Munyikwa and Suraj Nair and Avanika Narayan and Deepak Narayanan and Ben Newman and Allen Nie and Juan Carlos Niebles and Hamed Nilforoshan and Julian Nyarko and Giray Ogut and Laurel Orr and Isabel Papadimitriou and Joon Sung Park and Chris Piech and Eva Portelance and Christopher Potts and Aditi Raghunathan and Rob Reich and Hongyu Ren and Frieda Rong and Yusuf Roohani and Camilo Ruiz and Jack Ryan and Christopher Ré and Dorsa Sadigh and Shiori Sagawa and Keshav Santhanam and Andy Shih and Krishnan Srinivasan and Alex Tamkin and Rohan Taori and Armin W. Thomas and Florian Tramèr and Rose E. Wang and William Wang and Bohan Wu and Jiajun Wu and Yuhuai Wu and Sang Michael Xie and Michihiro Yasunaga and Jiaxuan You and Matei Zaharia and Michael Zhang and Tianyi Zhang and Xikun Zhang and Yuhui Zhang and Lucia Zheng and Kaitlyn Zhou and Percy Liang},
    year={2022},
    eprint={2108.07258},
    archivePrefix={arXiv},
    primaryClass={cs.LG},
    url={https://arxiv.org/abs/2108.07258}, 
}

@misc{llama,
    title={LLaMA: Open and Efficient Foundation Language Models}, 
    author={Hugo Touvron and Thibaut Lavril and Gautier Izacard and Xavier Martinet and Marie-Anne Lachaux and Timothée Lacroix and Baptiste Rozière and Naman Goyal and Eric Hambro and Faisal Azhar and Aurelien Rodriguez and Armand Joulin and Edouard Grave and Guillaume Lample},
    year={2023},
    eprint={2302.13971},
    archivePrefix={arXiv},
    primaryClass={cs.CL},
    url={https://arxiv.org/abs/2302.13971}, 
}

@misc{qwen3,
    title={Qwen3 Technical Report}, 
    author={An Yang and Anfeng Li and Baosong Yang and Beichen Zhang and Binyuan Hui and Bo Zheng and Bowen Yu and Chang Gao and Chengen Huang and Chenxu Lv and Chujie Zheng and Dayiheng Liu and Fan Zhou and Fei Huang and Feng Hu and Hao Ge and Haoran Wei and Huan Lin and Jialong Tang and Jian Yang and Jianhong Tu and Jianwei Zhang and Jianxin Yang and Jiaxi Yang and Jing Zhou and Jingren Zhou and Junyang Lin and Kai Dang and Keqin Bao and Kexin Yang and Le Yu and Lianghao Deng and Mei Li and Mingfeng Xue and Mingze Li and Pei Zhang and Peng Wang and Qin Zhu and Rui Men and Ruize Gao and Shixuan Liu and Shuang Luo and Tianhao Li and Tianyi Tang and Wenbiao Yin and Xingzhang Ren and Xinyu Wang and Xinyu Zhang and Xuancheng Ren and Yang Fan and Yang Su and Yichang Zhang and Yinger Zhang and Yu Wan and Yuqiong Liu and Zekun Wang and Zeyu Cui and Zhenru Zhang and Zhipeng Zhou and Zihan Qiu},
    year={2025},
    eprint={2505.09388},
    archivePrefix={arXiv},
    primaryClass={cs.CL},
    url={https://arxiv.org/abs/2505.09388}, 
}

@misc{gpt4,
    title={GPT-4 Technical Report}, 
    author={OpenAI and Josh Achiam and Steven Adler and Sandhini Agarwal and Lama Ahmad and Ilge Akkaya and Florencia Leoni Aleman and Diogo Almeida and Janko Altenschmidt and Sam Altman and Shyamal Anadkat and Red Avila and Igor Babuschkin and Suchir Balaji and Valerie Balcom and Paul Baltescu and Haiming Bao and Mohammad Bavarian and Jeff Belgum and Irwan Bello and Jake Berdine and Gabriel Bernadett-Shapiro and Christopher Berner and Lenny Bogdonoff and Oleg Boiko and Madelaine Boyd and Anna-Luisa Brakman and Greg Brockman and Tim Brooks and Miles Brundage and Kevin Button and Trevor Cai and Rosie Campbell and Andrew Cann and Brittany Carey and Chelsea Carlson and Rory Carmichael and Brooke Chan and Che Chang and Fotis Chantzis and Derek Chen and Sully Chen and Ruby Chen and Jason Chen and Mark Chen and Ben Chess and Chester Cho and Casey Chu and Hyung Won Chung and Dave Cummings and Jeremiah Currier and Yunxing Dai and Cory Decareaux and Thomas Degry and Noah Deutsch and Damien Deville and Arka Dhar and David Dohan and Steve Dowling and Sheila Dunning and Adrien Ecoffet and Atty Eleti and Tyna Eloundou and David Farhi and Liam Fedus and Niko Felix and Simón Posada Fishman and Juston Forte and Isabella Fulford and Leo Gao and Elie Georges and Christian Gibson and Vik Goel and Tarun Gogineni and Gabriel Goh and Rapha Gontijo-Lopes and Jonathan Gordon and Morgan Grafstein and Scott Gray and Ryan Greene and Joshua Gross and Shixiang Shane Gu and Yufei Guo and Chris Hallacy and Jesse Han and Jeff Harris and Yuchen He and Mike Heaton and Johannes Heidecke and Chris Hesse and Alan Hickey and Wade Hickey and Peter Hoeschele and Brandon Houghton and Kenny Hsu and Shengli Hu and Xin Hu and Joost Huizinga and Shantanu Jain and Shawn Jain and Joanne Jang and Angela Jiang and Roger Jiang and Haozhun Jin and Denny Jin and Shino Jomoto and Billie Jonn and Heewoo Jun and Tomer Kaftan and Łukasz Kaiser and Ali Kamali and Ingmar Kanitscheider and Nitish Shirish Keskar and Tabarak Khan and Logan Kilpatrick and Jong Wook Kim and Christina Kim and Yongjik Kim and Jan Hendrik Kirchner and Jamie Kiros and Matt Knight and Daniel Kokotajlo and Łukasz Kondraciuk and Andrew Kondrich and Aris Konstantinidis and Kyle Kosic and Gretchen Krueger and Vishal Kuo and Michael Lampe and Ikai Lan and Teddy Lee and Jan Leike and Jade Leung and Daniel Levy and Chak Ming Li and Rachel Lim and Molly Lin and Stephanie Lin and Mateusz Litwin and Theresa Lopez and Ryan Lowe and Patricia Lue and Anna Makanju and Kim Malfacini and Sam Manning and Todor Markov and Yaniv Markovski and Bianca Martin and Katie Mayer and Andrew Mayne and Bob McGrew and Scott Mayer McKinney and Christine McLeavey and Paul McMillan and Jake McNeil and David Medina and Aalok Mehta and Jacob Menick and Luke Metz and Andrey Mishchenko and Pamela Mishkin and Vinnie Monaco and Evan Morikawa and Daniel Mossing and Tong Mu and Mira Murati and Oleg Murk and David Mély and Ashvin Nair and Reiichiro Nakano and Rajeev Nayak and Arvind Neelakantan and Richard Ngo and Hyeonwoo Noh and Long Ouyang and Cullen O'Keefe and Jakub Pachocki and Alex Paino and Joe Palermo and Ashley Pantuliano and Giambattista Parascandolo and Joel Parish and Emy Parparita and Alex Passos and Mikhail Pavlov and Andrew Peng and Adam Perelman and Filipe de Avila Belbute Peres and Michael Petrov and Henrique Ponde de Oliveira Pinto and Michael and Pokorny and Michelle Pokrass and Vitchyr H. Pong and Tolly Powell and Alethea Power and Boris Power and Elizabeth Proehl and Raul Puri and Alec Radford and Jack Rae and Aditya Ramesh and Cameron Raymond and Francis Real and Kendra Rimbach and Carl Ross and Bob Rotsted and Henri Roussez and Nick Ryder and Mario Saltarelli and Ted Sanders and Shibani Santurkar and Girish Sastry and Heather Schmidt and David Schnurr and John Schulman and Daniel Selsam and Kyla Sheppard and Toki Sherbakov and Jessica Shieh and Sarah Shoker and Pranav Shyam and Szymon Sidor and Eric Sigler and Maddie Simens and Jordan Sitkin and Katarina Slama and Ian Sohl and Benjamin Sokolowsky and Yang Song and Natalie Staudacher and Felipe Petroski Such and Natalie Summers and Ilya Sutskever and Jie Tang and Nikolas Tezak and Madeleine B. Thompson and Phil Tillet and Amin Tootoonchian and Elizabeth Tseng and Preston Tuggle and Nick Turley and Jerry Tworek and Juan Felipe Cerón Uribe and Andrea Vallone and Arun Vijayvergiya and Chelsea Voss and Carroll Wainwright and Justin Jay Wang and Alvin Wang and Ben Wang and Jonathan Ward and Jason Wei and CJ Weinmann and Akila Welihinda and Peter Welinder and Jiayi Weng and Lilian Weng and Matt Wiethoff and Dave Willner and Clemens Winter and Samuel Wolrich and Hannah Wong and Lauren Workman and Sherwin Wu and Jeff Wu and Michael Wu and Kai Xiao and Tao Xu and Sarah Yoo and Kevin Yu and Qiming Yuan and Wojciech Zaremba and Rowan Zellers and Chong Zhang and Marvin Zhang and Shengjia Zhao and Tianhao Zheng and Juntang Zhuang and William Zhuk and Barret Zoph},
    year={2024},
    eprint={2303.08774},
    archivePrefix={arXiv},
    primaryClass={cs.CL},
    url={https://arxiv.org/abs/2303.08774}, 
}

@INPROCEEDINGS{maskedAutoEncoders,
  author={He, Kaiming and Chen, Xinlei and Xie, Saining and Li, Yanghao and Dollár, Piotr and Girshick, Ross},
  booktitle={2022 IEEE/CVF Conference on Computer Vision and Pattern Recognition (CVPR)}, 
  title={Masked Autoencoders Are Scalable Vision Learners}, 
  year={2022},
  volume={},
  number={},
  pages={15979-15988},
  doi={10.1109/CVPR52688.2022.01553}}

@InProceedings{swin,
    author    = {Liu, Ze and Lin, Yutong and Cao, Yue and Hu, Han and Wei, Yixuan and Zhang, Zheng and Lin, Stephen and Guo, Baining},
    title     = {Swin Transformer: Hierarchical Vision Transformer Using Shifted Windows},
    booktitle = {Proceedings of the IEEE/CVF International Conference on Computer Vision (ICCV)},
    month     = {October},
    year      = {2021},
    pages     = {10012-10022}
}

@INPROCEEDINGS{residual_learning,
  author={He, Kaiming and Zhang, Xiangyu and Ren, Shaoqing and Sun, Jian},
  booktitle={2016 IEEE Conference on Computer Vision and Pattern Recognition (CVPR)}, 
  title={Deep Residual Learning for Image Recognition}, 
  year={2016},
  volume={},
  number={},
  pages={770-778},
  doi={10.1109/CVPR.2016.90}}

@inproceedings{shifman2024losttranslationmodernneural,
  title={Lost in translation: modern neural networks still struggle with small realistic image transformations},
  author={Shifman, Ofir and Weiss, Yair},
  booktitle={European Conference on Computer Vision},
  pages={231--247},
  year={2024},
  organization={Springer}
}

@Article{electronics14091882,
AUTHOR = {Kim, Kana and Kakani, Vijay and Kim, Hakil},
TITLE = {Automatic Pruning and Quality Assurance of Object Detection Datasets for Autonomous Driving},
JOURNAL = {Electronics},
VOLUME = {14},
YEAR = {2025},
NUMBER = {9},
ARTICLE-NUMBER = {1882},
URL = {https://www.mdpi.com/2079-9292/14/9/1882},
ISSN = {2079-9292},
DOI = {10.3390/electronics14091882}
}

@phdthesis{Hwang2023Improvement,
  title     = {Improvement of Object Detection in Autonomous Driving using Communication Systems},
  author    = {Hwang, Sunwook},
  school    = {Seoul National University},
  year      = {2023},
  type      = {Ph.D. dissertation},
  address   = {Seoul},
  month     = {aug},
  language  = {eng},
  url       = {https://hdl.handle.net/10371/196434},
  note      = {Supervised by Se-Woong Park},
  pages     = {x, 83}
}

@InProceedings{Yin_2024_CVPR,
    author    = {Yin, Yingda and Liu, Yuzheng and Xiao, Yang and Cohen-Or, Daniel and Huang, Jingwei and Chen, Baoquan},
    title     = {SAI3D: Segment Any Instance in 3D Scenes},
    booktitle = {Proceedings of the IEEE/CVF Conference on Computer Vision and Pattern Recognition (CVPR)},
    month     = {June},
    year      = {2024},
    pages     = {3292-3302}
}

@article{Everingham2010Jun,
	author = {Everingham, Mark and Van Gool, Luc and Williams, Christopher K. I. and Winn, John and Zisserman, Andrew},
	title = {{The Pascal Visual Object Classes (VOC) Challenge}},
	journal = {Int. J. Comput. Vision},
	volume = {88},
	number = {2},
	pages = {303--338},
	year = {2010},
	month = jun,
	issn = {1573-1405},
	publisher = {Springer US},
	doi = {10.1007/s11263-009-0275-4}
}

@article{pariza2024near,
  title={Near, far: Patch-ordering enhances vision foundation models' scene understanding},
  author={Pariza, Valentinos and Salehi, Mohammadreza and Burghouts, Gertjan and Locatello, Francesco and Asano, Yuki M},
  journal={arXiv preprint arXiv:2408.11054},
  year={2024}
}

@article{salehi2025mosic,
  title={MoSiC: Optimal-Transport Motion Trajectory for Dense Self-Supervised Learning},
  author={Salehi, Mohammadreza and Venkataramanan, Shashanka and Simion, Ioana and Gavves, Efstratios and Snoek, Cees GM and Asano, Yuki M},
  journal={arXiv preprint arXiv:2506.08694},
  year={2025}
}

@inproceedings{salehi2023time,
  title={Time does tell: Self-supervised time-tuning of dense image representations},
  author={Salehi, Mohammadreza and Gavves, Efstratios and Snoek, Cees GM and Asano, Yuki M},
  booktitle={Proceedings of the IEEE/CVF International Conference on Computer Vision},
  pages={16536--16547},
  year={2023}
}

@inproceedings{ziegler2022self,
  title={Self-supervised learning of object parts for semantic segmentation},
  author={Ziegler, Adrian and Asano, Yuki M},
  booktitle={Proceedings of the IEEE/CVF conference on computer vision and pattern recognition},
  pages={14502--14511},
  year={2022}
}

@article{johnson2019billion,
  title={Billion-scale similarity search with {GPUs}},
  author={Johnson, Jeff and Douze, Matthijs and J{\'e}gou, Herv{\'e}},
  journal={IEEE Transactions on Big Data},
  volume={7},
  number={3},
  pages={535--547},
  year={2019},
  publisher={IEEE}
}

@misc{simeoni2025dinov3,
  title={{DINOv3}},
  author={Sim{\'e}oni, Oriane and Vo, Huy V. and Seitzer, Maximilian and Baldassarre, Federico and Oquab, Maxime and Jose, Cijo and Khalidov, Vasil and Szafraniec, Marc and Yi, Seungeun and Ramamonjisoa, Micha{\"e}l and Massa, Francisco and Haziza, Daniel and Wehrstedt, Luca and Wang, Jianyuan and Darcet, Timoth{\'e}e and Moutakanni, Th{\'e}o and Sentana, Leonel and Roberts, Claire and Vedaldi, Andrea and Tolan, Jamie and Brandt, John and Couprie, Camille and Mairal, Julien and J{\'e}gou, Herv{\'e} and Labatut, Patrick and Bojanowski, Piotr},
  year={2025},
  eprint={2508.10104},
  archivePrefix={arXiv},
  primaryClass={cs.CV},
  url={https://arxiv.org/abs/2508.10104},
}

\newpage
\appendix
\section{Licenses for assets}
\label{app:licenses}

All datasets and models used in this work are released under open licenses. 
We list them in \autoref{tab:license-info} below.



\begin{table}[H]
  \centering
  \footnotesize
  \caption{\textbf{Licenses for datasets and models}. All assets are used in compliance with their respective licenses.}
  \begin{tabular}{ll}
    \toprule
    \textbf{Asset} & \textbf{License} \\
    \midrule
    MVImgNet & CC BY-NC 4.0 (code), dataset Terms of Use \\
    Hummingbird & MIT License \\
    CLIP & MIT \\
    DINO & Apache 2.0 \\
    DINOv2 & Apache 2.0 \\
    DINOv3 & DINOv3 License \\
    SigLIP2 & Apache 2.0 \\
    C-RADIOv2 & NVIDIA Open Model License \\
    TIPS & Apache 2.0 (code), CC BY 4.0 (docs) \\
    \bottomrule
  \end{tabular}
  \label{tab:license-info}
\end{table}

\section{The MVImgNet dataset} \label{App:MVImgNet}

\begin{figure}[H] 
    \centering
    \includegraphics[width=\linewidth]{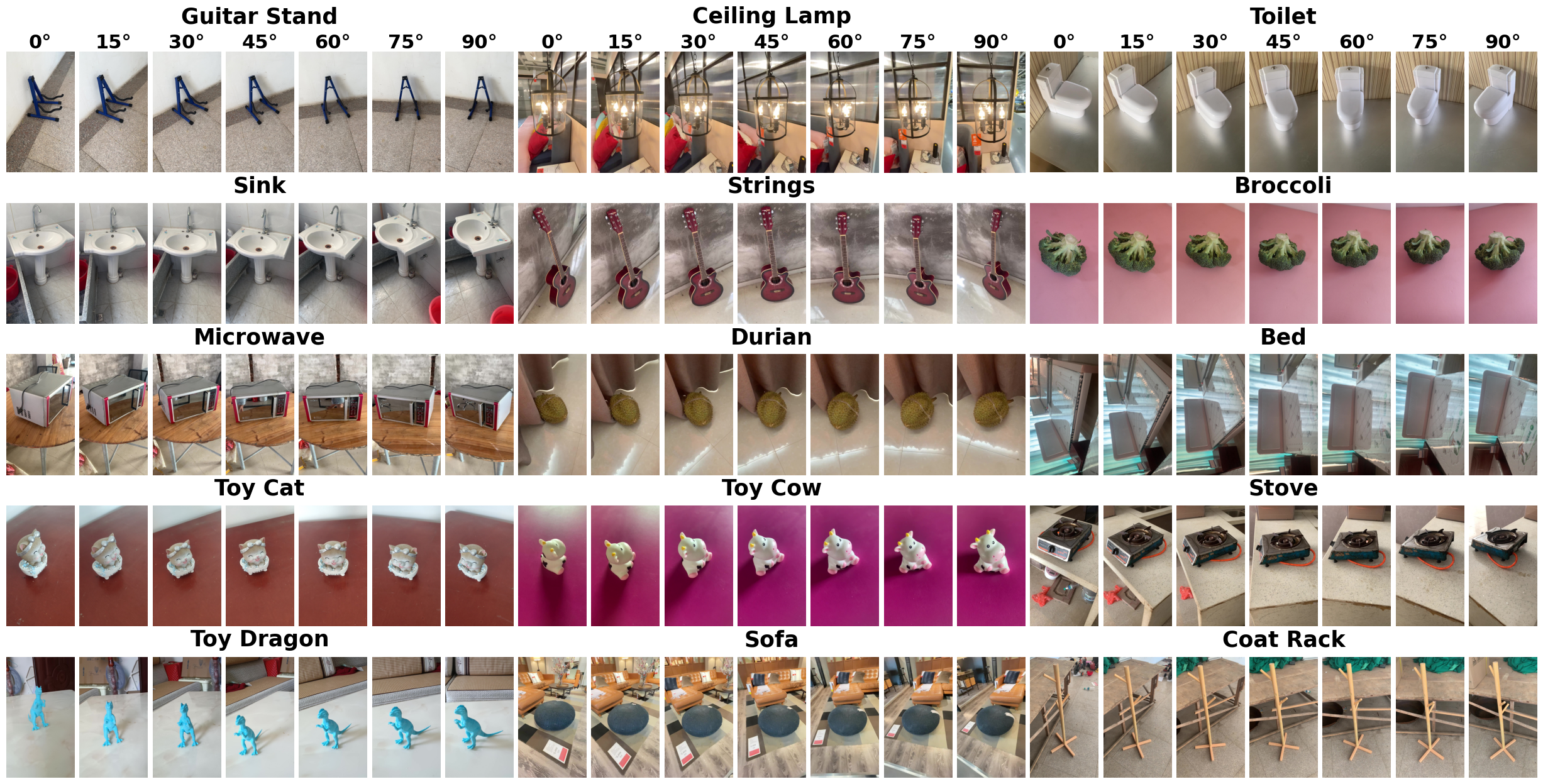}
    \caption{\textbf{Multi-view categories}. A representative instance is shown for each of the 15 selected MVImgNet classes across all viewpoint bins (0\(^\circ\)--90\(^\circ\)).
    Based on the category taxonomy of MVImgNet \cite{Yu_2023_CVPR_mvimgnet}, we have examples from both the \textit{food} and \textit{artifacts} parent classes, which encompass the following child classes: \textit{fruits and vegetables}, \textit{toy dolls}, and a series of \textit{instruments}: \textit{furniture}, \textit{kitchen ware}, \textit{home appliances}, and \textit{entertainment}.}
    \label{fig:all_angles_objects}
\end{figure}

The statistics for the reorganized subset of MVImgNet~\cite{Yu_2023_CVPR_mvimgnet} we used are shown in~\autoref{tab:angle-selection-stats}. The selected subset includes 15 object classes, each represented across 7 standardized angle bins: 0\(^\circ\), 15\(^\circ\), 30\(^\circ\), 45\(^\circ\), 60\(^\circ\), 75\(^\circ\), and 90\(^\circ\).
\autoref{fig:all_angles_objects} shows an instance of each angle for each object class.

\begin{table}[H]
    \caption{\textbf{Angle selection accuracy per class}. Mean and standard deviation of angular error (in degrees) for each object class after selecting views closest to the predefined angle bins.
    }
    \centering
    \footnotesize
    \begin{tabular}{clccc}
        \toprule
        \textbf{Class number} & \textbf{Category} & \textbf{Std. error} & \textbf{Mean error} & \textbf{Images per bin} \\
        \midrule
        7   & Stove        & 1.28 & 0.01  & 197 \\
        8   & Sofa         & 1.15 & -0.04 & 91 \\
        19  & Microwave    & 1.44 & -0.04 & 120 \\
        46  & Bed          & 1.17 & -0.00 & 23 \\
        57  & Toy cat      & 1.96 & -0.01 & 783 \\
        60  & Toy cow      & 1.96 & -0.02 & 735 \\
        70  & Toy dragon   & 1.62 & 0.01  & 627 \\
        99  & Coat rack    & 1.41 & -0.02 & 97 \\
        100 & Guitar stand & 1.53 & 0.04  & 218 \\
        113 & Ceiling lamp & 1.64 & -0.03 & 154 \\
        125 & Toilet       & 1.31 & 0.02  & 58 \\
        126 & Sink         & 1.20 & -0.12 & 30 \\
        152 & Strings      & 1.25 & 0.03  & 192 \\
        166 & Broccoli     & 2.04 & -0.03 & 210 \\
        196 & Durian       & 1.65 & 0.03  & 758 \\
        \bottomrule
    \end{tabular}
    \label{tab:angle-selection-stats}
\end{table}

\newpage

\section{Model specifics}
\label{App:Model_Specifics}

The following table details the ViT-based encoders used in our experiments, including their patch configurations and feature dimensionality.

\begin{table}[H]
    \centering
    \footnotesize
    \caption{\textbf{Vision Transformer configurations}. We show the architectural and input settings of the models evaluated in our experiments. Input, patch, and batch sizes are indicated with Input sz., Patch sz., and Batch sz., respectively.}
    \label{tab:model-specs}
    \begin{tabular}{llccccc}
        \toprule
        \textbf{Model} & \textbf{Architecture} & \textbf{Input sz.} & \textbf{Patch sz.} & \textbf{\# patches} & \textbf{Feat. dim.} & \textbf{Batch sz.}\\
        \midrule
        DINO & ViT-B/16               & 512 & 16 & 1024 & 768 & 4 \\
        DINOv2 & ViT-B/14             & 504 & 14 & 1296 & 768 & 4 \\
        DINOv3 & ViT-B/16             & 512 & 16 & 1024 & 768 & 4 \\
        OpenAI CLIP & ViT-B/16        & 512 & 16 & 1024 & 768 & 4 \\
        C-RADIOv2 & ViT-B/16-CPE      & 512 & 16 & 1024 & 768 & 4 \\
        SigLIP2 & ViT-B/16-512             & 512 & 16 & 1024 & 768 & 4 \\
        TIPS & ViT-B/14-HR       & 504 & 14 & 1296 & 768 & 4 \\ 
        \bottomrule
    \end{tabular}
\end{table}

The models differ in their native input resolutions:
\(224 \times 224\) for CLIP ViT-B/16~\cite{CLIP}, DINO ViT-B/16~\cite{DINO}, DINOv3 ViT-B/16~\cite{simeoni2025dinov3}
and C-RADIOv2 ViT-B/16-CPE~\cite{radiov2.5,am-radio};
\(512 \times 512\) for DINOv2 ViT-B/14~\cite{DINOv2};
\(504 \times 504\) for SigLIP2 B/16-512~\cite{SIGLIPv2};
and \(448 \times 448\) for TIPS ViT-B/14-HR~\cite{tips_paper}.

\section{Computational resources}
\label{App:Computational_Resources}

All experiments were executed on the Snellius HPC cluster to ensure high-performance processing and reproducibility. We ran our experiments on a single A100 node consisting of 4 $\times$ NVIDIA A100 GPUs and 72 CPU cores, with a maximum wall time of 30 hours allocated per job to account for memory and processing constraints. 

Regarding software and environment settings, the search index was distributed using FAISS sharding across all available GPUs. We utilized a batch size of 4 for all runs. To ensure experimental consistency, a global random seed of 42 was set across Python, NumPy, PyTorch, and CUDA.

\section{Reproduction results}
\label{app:reproduction}

\begin{table}[H]
\centering
\footnotesize
\caption{\textbf{Reproduction results}. We report accuracy (\%) at different evaluation scales. 
``Reported'' indicates values taken from prior published results, while ``Reproduced'' refers to our reproduction with a batch size (BS). 
Bold values highlight the best result for each model-scale configuration. Overall, our reproduced scores are close to or exceed the reported results in the original paper.
}

\begin{tabular}{llccc}
    \toprule
    \textbf{Model} & \textbf{Source} &
    \boldmath{\(1024\times10^2\)} &
    \boldmath{\(1024\times10^3\)} &
    \boldmath{\(1024\times10^4\)} \\
    \midrule
    ViT-S/16 & Reproduced, BS: 256     & 37.5 & 45.0 & -- \\
             & Reproduced, BS: 64      & \textbf{37.9} & \textbf{45.1} & \textbf{49.3} \\
             & Reported, BS: 64   & 37.2 & 43.1 & 46.6 \\
    \midrule
    ViT-B/16 & Reproduced, BS: 64      & \textbf{48.0} & \textbf{54.7} & -- \\
             & Reproduced, BS: 32      & 47.8 & 54.6 & -- \\
             & Reproduced, BS: 8       & --    & --    & \textbf{57.9} \\
             & Reported, BS: 64   & 44.9 & 50.8 & 55.7 \\
    \midrule
    ViT-S/14 & Reproduced, BS: 64      & 69.6 & \textbf{75.1} & \textbf{77.0} \\
             & Reported, BS: 64   & \textbf{70.2} & 74.9 & 77.0 \\
    \midrule
    ViT-B/14 & Reproduced, BS: 64      & 68.0 & 74.0 & -- \\
             & Reproduced, BS: 8       & --    & --    & \textbf{76.6} \\
             & Reported, BS: 64   & \textbf{69.1} & \textbf{74.6} & 76.9 \\
    \midrule
    ViT-L/14 & Reproduced, BS: 64      & 64.1 & 71.2 & -- \\
             & Reproduced, BS: 8       & --    & --    & 74.4 \\
             & Reported, BS: 64   & \textbf{64.6} & \textbf{71.7} & \textbf{74.8} \\
    \midrule
    ViT-G/14 & Reproduced, BS: 32      & \textbf{62.4} & --    & -- \\
             & Reproduced, BS: 16      & --    & \textbf{70.1} & -- \\
             & Reproduced, BS: 8       & --    & --    & 73.3 \\
             & Reported, BS: 64   & 62.3 & 69.9 & \textbf{73.6} \\
    \bottomrule
\end{tabular}
\label{tab:hbird-reproduction}
\end{table}

\FloatBarrier
\newpage
\section{Additional results}
\label{app:additional_results}

\subsection{Experiment A}
\label{app:expa}
In experiment A, we assess model generalization across view-angle difficulties using per-class performance curves. In the figures below, the bars represent mean mIoU for unseen validation bins, and dots mark reference viewpoints. Performance typically declines as viewpoint distance increases from the reference bins, consistent with the trend in \autoref{fig:expa_all_classes}. 
Across viewpoint difficulties, DINOv3, DINOv2, DINO, and CLIP consistently outperform all other models. C-RADIOv2 follows closely, while SigLIP2 and especially TIPS degrade substantially under extreme viewpoints (with TIPS showing the weakest overall robustness). DINOv3 generally outperforms all models; however, this does not hold for the \textit{coat rack} and \textit{guitar stand} classes, where, in the Extreme difficulty scenario, it becomes one of the worst-performing models.

Across all models, the background class performs best; other classes that perform well are \textit{durian}, \textit{broccoli}, and \textit{strings}. These tend to generalize above average under unseen viewpoints. In contrast, \textit{bed}, \textit{coat rack}, \textit{guitar stand}, and \textit{sink} perform worst (the largest negative generalization gaps). For the \textit{sink} class, DINOv3 significantly outperforms all other models. For \textit{bed}, the drop is explained by poor ground-truth annotations that likely reduce model scores (as explored in Appendix~\ref{app:bed_gt_analysis}).

\begin{figure}[H]
    \centering
    \begin{minipage}[t]{0.49\linewidth}
        \centering
        \includegraphics[width=\linewidth]{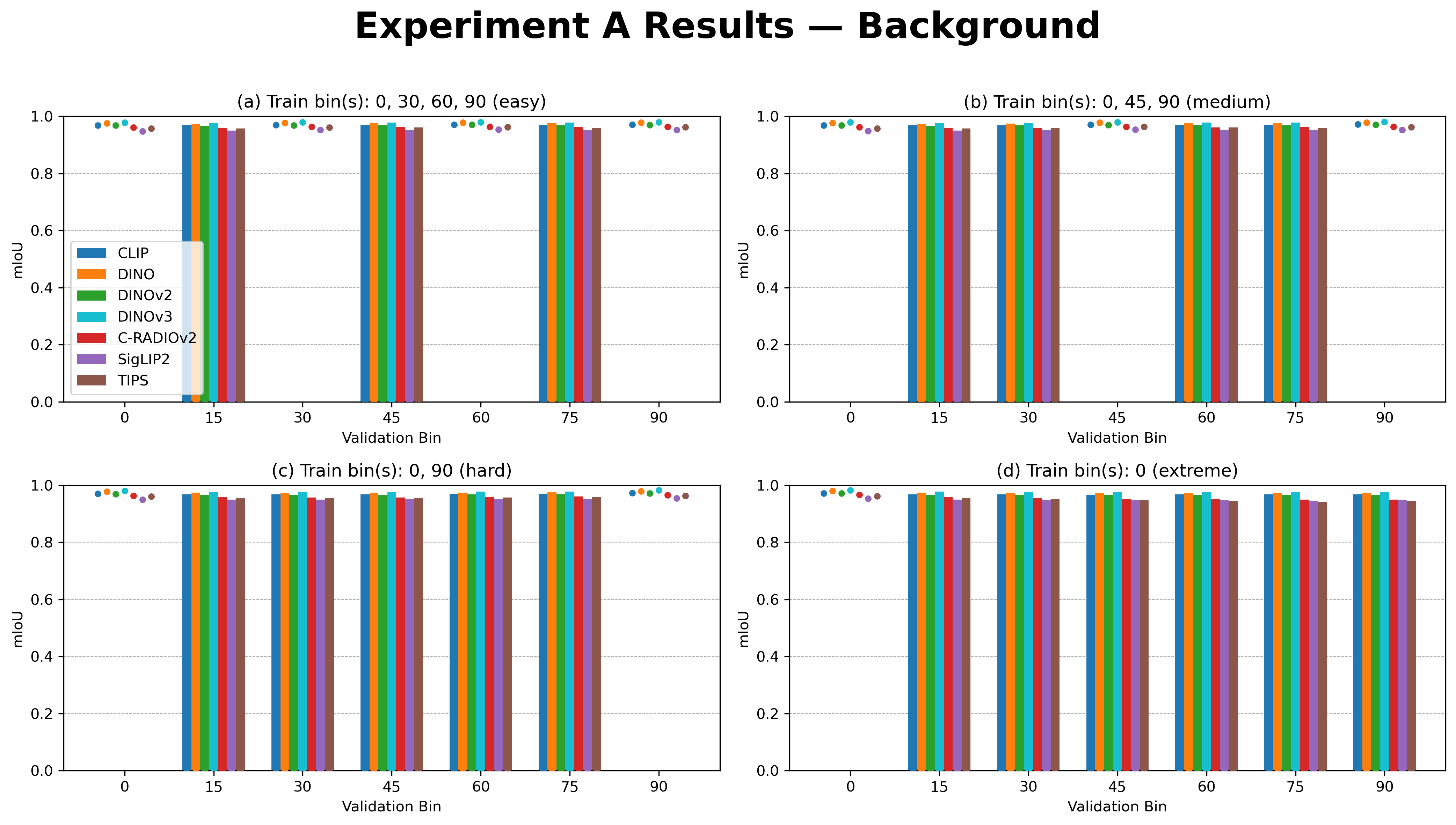}
        \caption{\textbf{Experiment A: background}. Viewpoint generalization for the background class across 4 training difficulty settings: (a) easy, (b) medium, (c) hard, and (d) extreme. All models achieve near-perfect consistency across bins, indicating that the background segmentation task remains invariant to viewpoint changes.  
        }
        \label{fig:expa_bg}
    \end{minipage}
    \hfill
    \begin{minipage}[t]{0.49\linewidth}
        \centering
        \includegraphics[width=\linewidth]{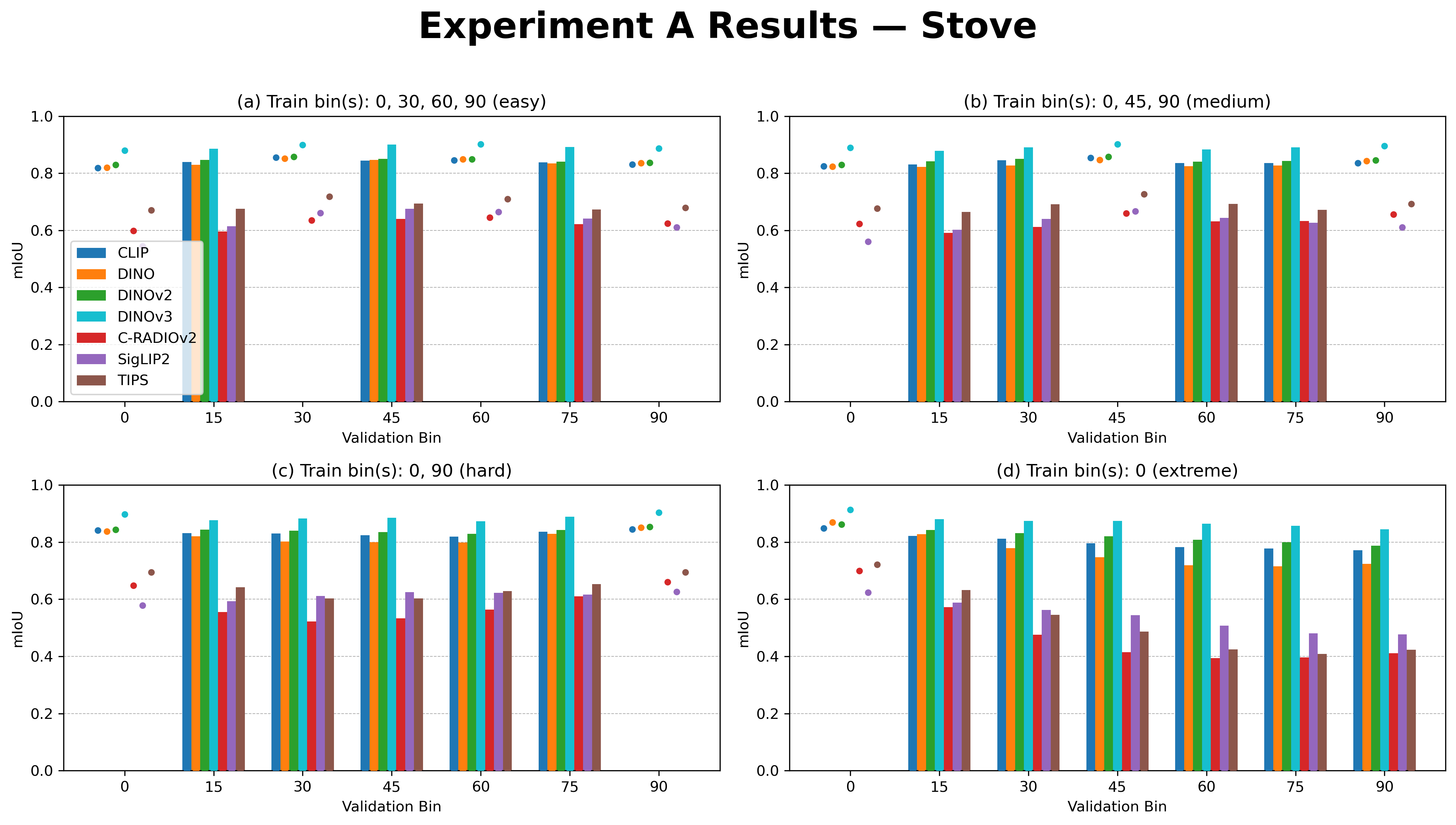}
        \caption{\textbf{Experiment A: stove}. The stove class perform slightly above the dataset average shown in in \autoref{fig:expa_all_classes}. The mIoU remains consistently high, and relative difference stays positive across all models. SigLIP2 and CLIP show the strongest viewpoint robustness. Degradation across validation bins follows the general pattern of \autoref{fig:expa_all_classes}, with no major model-specific failures.}
        \label{fig:expa_stove}
    \end{minipage}
    
\end{figure}

\FloatBarrier
\newpage

\begin{figure}[H]
    \centering
    \begin{minipage}[t]{0.49\linewidth}
        \centering
        \includegraphics[width=\linewidth]{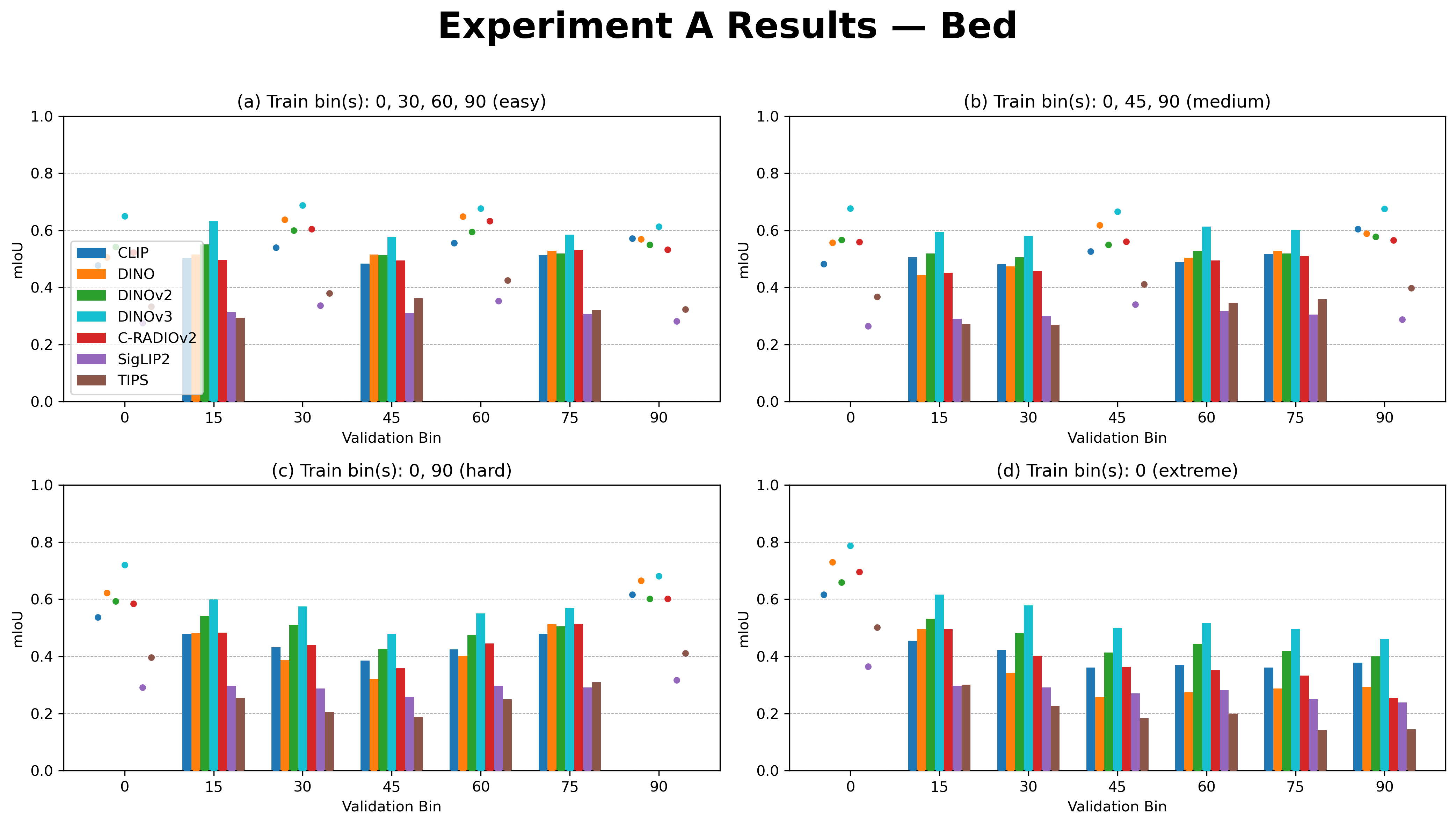}
        \caption{\textbf{Experiment A: bed}. The bed class scores far below the all-class baseline, with it's mIoU scores remaining low even for strong models. This degradation is largely explained by systematically incorrect ground-truth masks, as seen in Figures~\ref{fig:bed}--\ref{fig:bed6} we analyze in Appendix~\ref{app:bed_gt_analysis}. The figure reflects this annotation-driven performance drop.}
        \label{fig:12_bed}
    \end{minipage}
    \hfill
    \begin{minipage}[t]{0.49\linewidth}
        \centering
        \includegraphics[width=\linewidth]{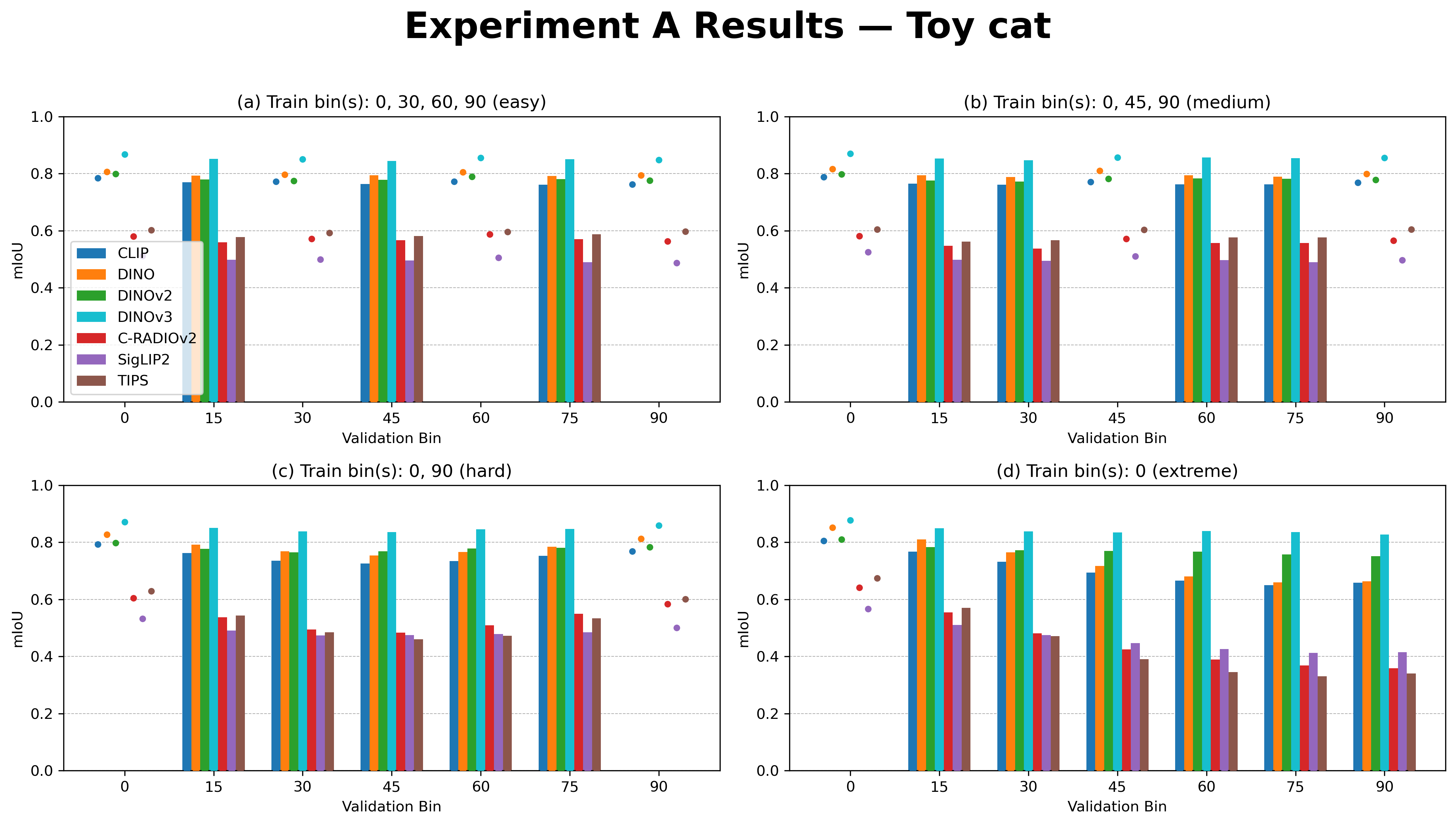}
        \caption{\textbf{Experiment A: toy cat}. The toy cat performance is very similar to the dataset-wide average observed in \autoref{fig:expa_all_classes}, with relative differences near zero. Strong models (CLIP, DINO v1-v3) achieve high mIoU (0.77--0.84), and accuracy drops consistently across difficulty settings.}
        \label{fig:expa_toycat}
    \end{minipage}
\end{figure}

\begin{figure}[H]
    \centering
    \begin{minipage}[t]{0.49\linewidth}
        \centering
        \includegraphics[width=\linewidth]{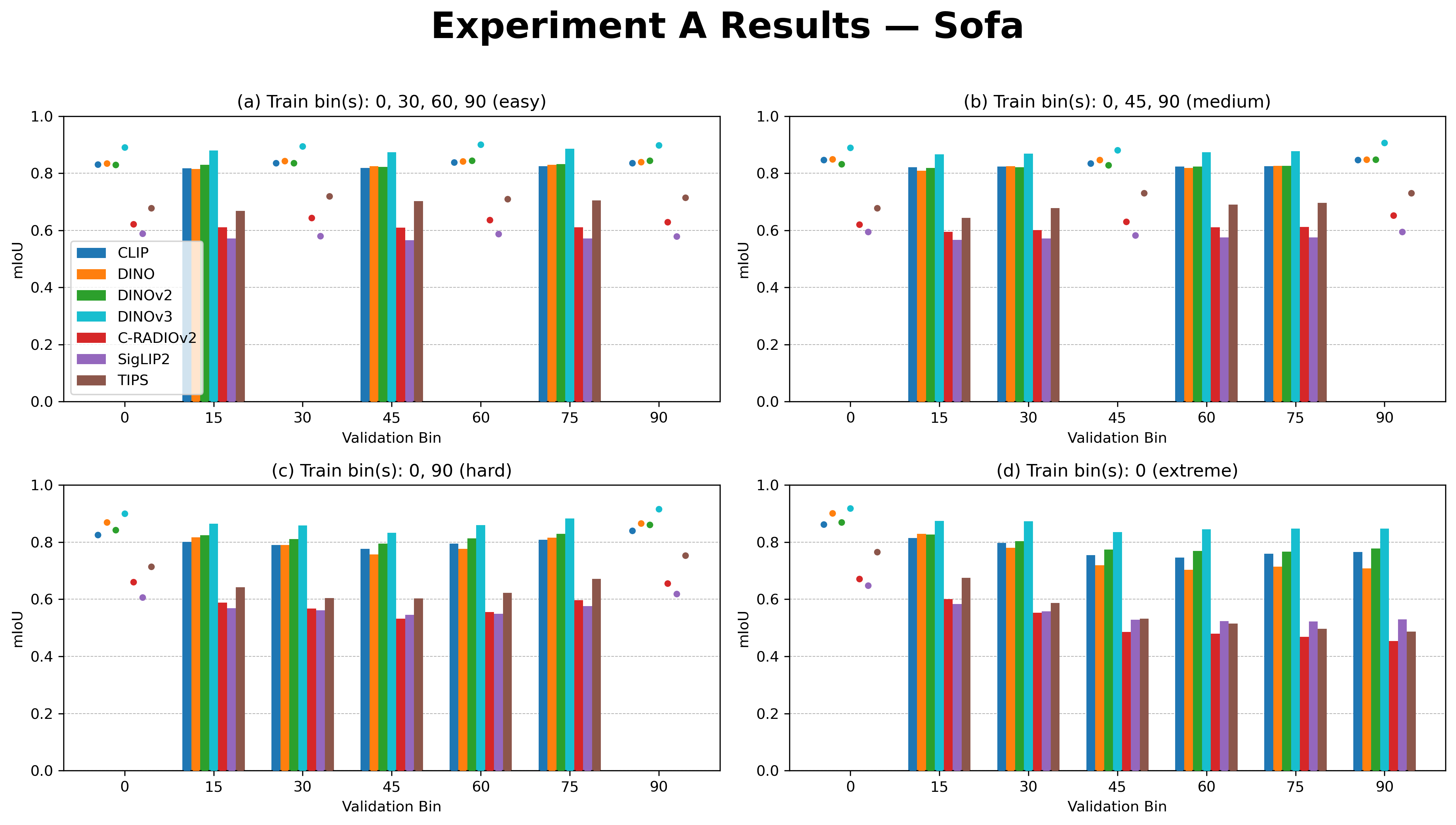}
        \caption{\textbf{Experiment A: sofa}. The sofa class performs slightly above the overall class average for every model and difficulty setting. 
        The mIoU performance decreases smoothly with increased viewpoint difficulty, matching the overall experiment A trend observed in~\autoref{fig:expa_all_classes}.}
        \label{fig:expa_sofa}
    \end{minipage}
    \hfill
    \begin{minipage}[t]{0.49\linewidth}
        \centering
        \includegraphics[width=\linewidth]{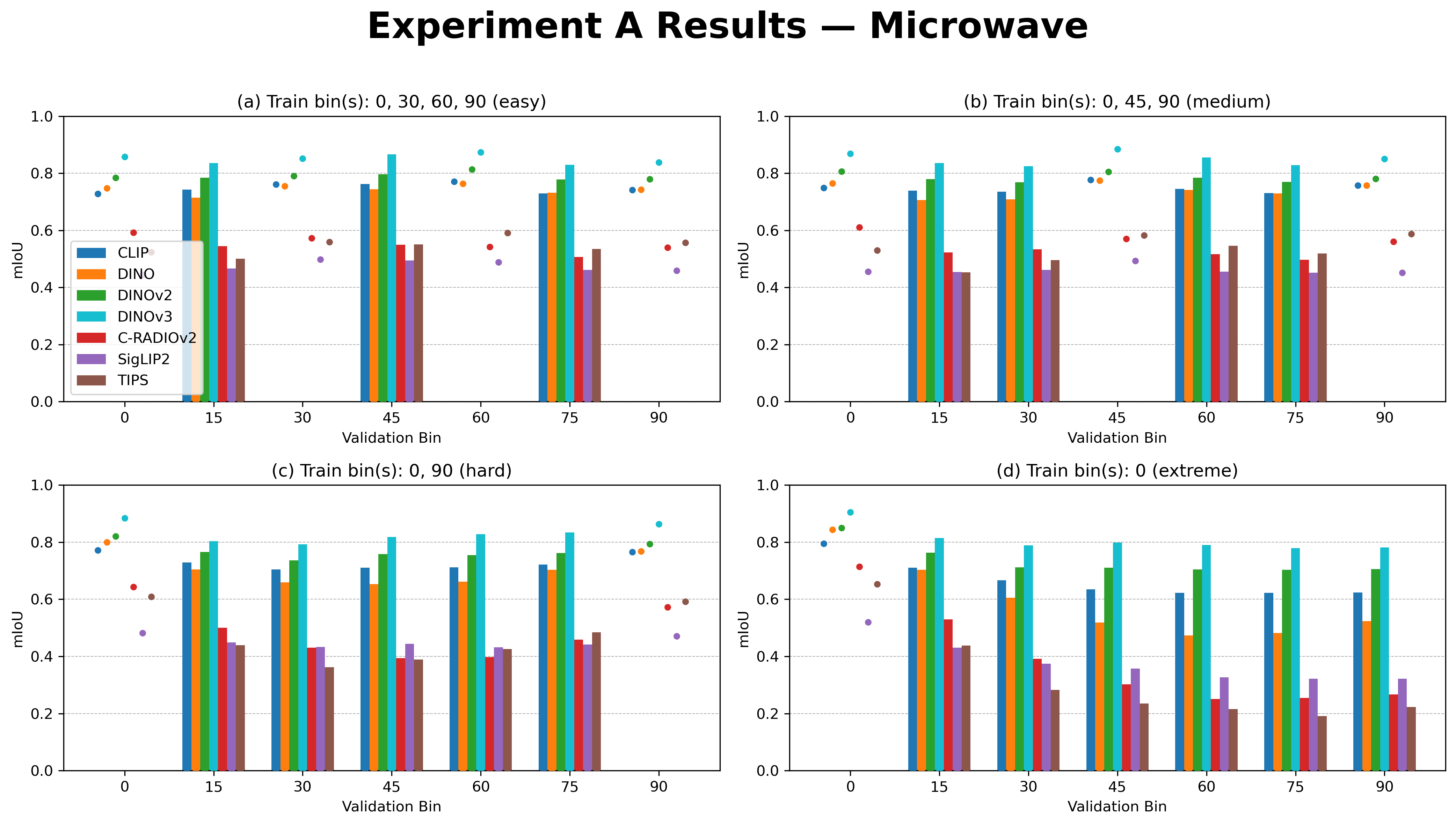}
        \caption{\textbf{Experiment A: microwave}. This class underperforms relative to~\autoref{fig:expa_all_classes}, with the largest drops appearing under Hard and Extreme difficulty. Model behavior diverges notably: DINOv3, DINOv2 and CLIP maintain comparatively strong performance, whereas DINO shows a mild decline. TIPS and SigLIP2 appear to struggle when segmenting the microwave category and degrade sharply under the Extreme condition, with a pronounced collapse beyond the 45\(^\circ\) bin. Despite these weaker generalization patterns, the overall degradation across viewpoint bins follows the expected trend.}
        \label{fig:expa_microwave}
    \end{minipage}
\end{figure}

\FloatBarrier
\newpage
\begin{figure}[H]
    \centering
    \begin{minipage}[t]{0.49\linewidth}
        \centering
        \includegraphics[width=\linewidth]{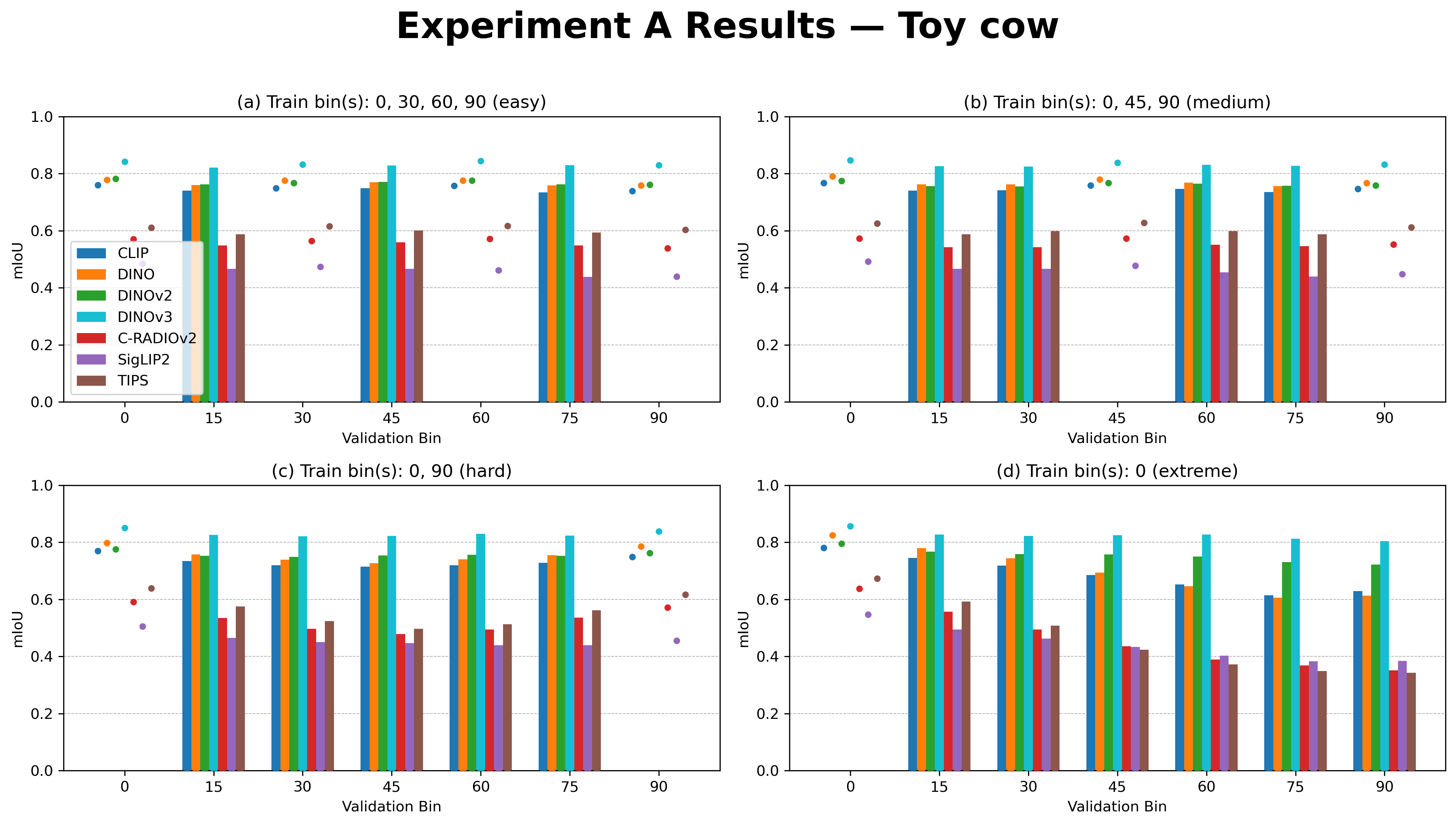}
        \caption{\textbf{Experiment A: toy cow}. The toy cow category shows slightly lower performance across all models on unseen bins, placing it among the weaker classes. Most noticeable is the lower performance of SIGLIP2 in the easy and medium task (see for example angle bin 75\(^\circ\) in the easy task).}
        \label{fig:expa_toy_cow}
    \end{minipage}
    \hfill
    \begin{minipage}[t]{0.49\linewidth}
        \centering
        \includegraphics[width=\linewidth]{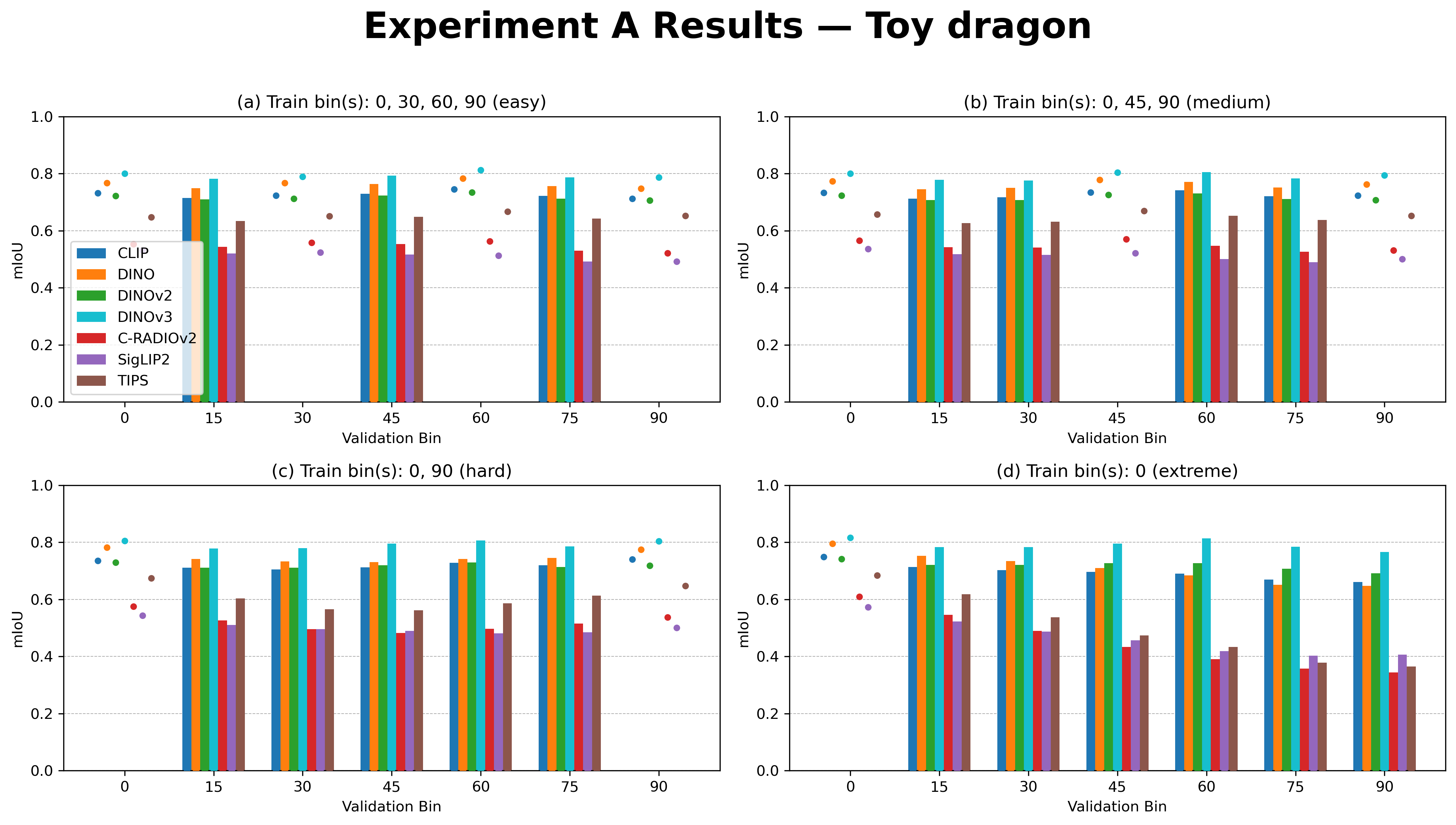}
        \caption{\textbf{Experiment A: toy dragon}. This class exhibits mildly negative gaps, in particular C-RADIOv2 is always lower than TIPS (when comparing to~\autoref{fig:expa_all_classes}). As a whole though, its performance trends mirror the general degradation observed of weaker models struggling as viewpoint distance increases.}
        \label{fig:expa_toy_dragon}
    \end{minipage}
\end{figure}

\begin{figure}[H]
    \centering
    \begin{minipage}[t]{0.49\linewidth}
        \centering
        \includegraphics[width=\linewidth]{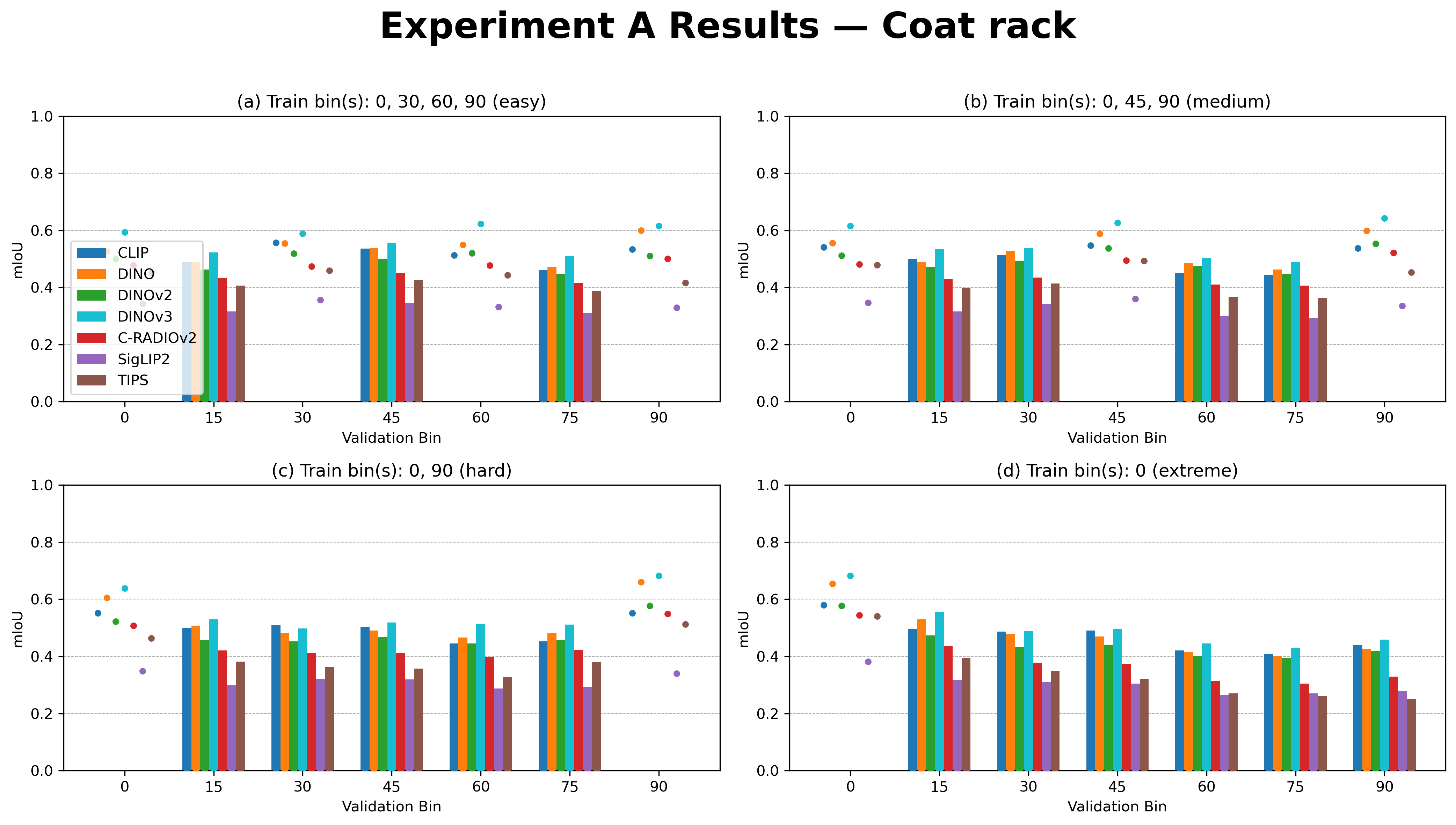}
        \caption{\textbf{Experiment A: coat rack}. 
        The coat rack is the worst-performing class overall, with mIoU remaining below 0.7 across all validation bins. 
        As seen in \autoref{fig:all_angles_objects}, the object’s structure is extremely thin, which likely makes it highly sensitive to viewpoint changes and contributes to its weak generalization. It remains an open question whether this difficulty is intrinsic to the geometry or arises from dataset-specific factors. This could be an area of further investigation.
        }
        \label{fig:expa_coat_rack}
    \end{minipage}
        \hfill
        \begin{minipage}[t]{0.49\linewidth}
        \centering
        \includegraphics[width=\linewidth]{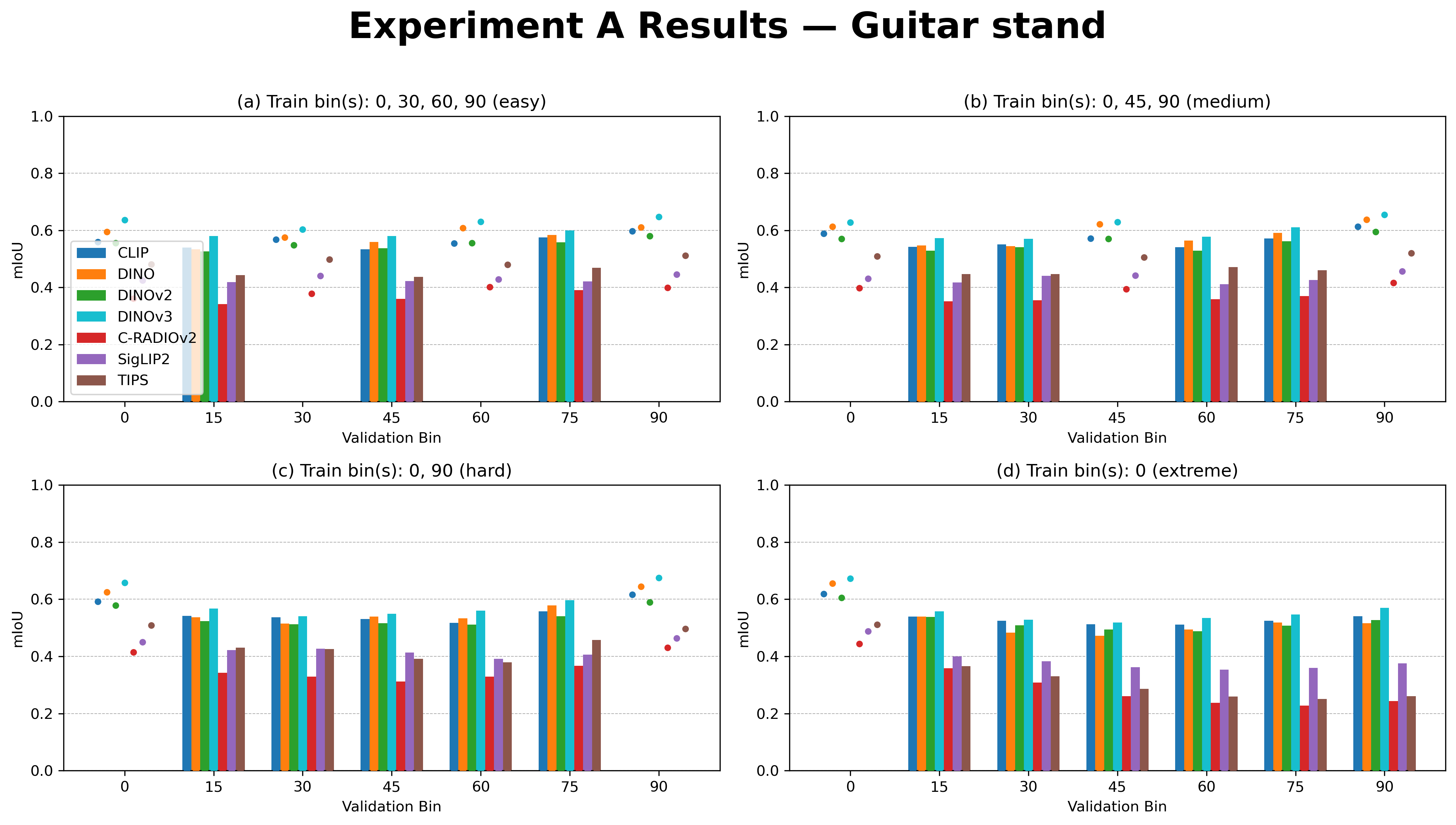}
        \caption{\textbf{Experiment A: guitar stand}. The Guitar stand class scores well below the baseline for all models and difficulties. 
        Like the coat rack, the thin structure likely contributes to poor performance.  
        However, the exact cause remains unclear and is a point for future investigation
        .}
        \label{fig:expa_guitar_stand}
    \end{minipage}
\end{figure}

\FloatBarrier
\newpage
\begin{figure}[H]
    \centering
    \begin{minipage}[t]{0.49\linewidth}
        \centering
        \includegraphics[width=\linewidth]{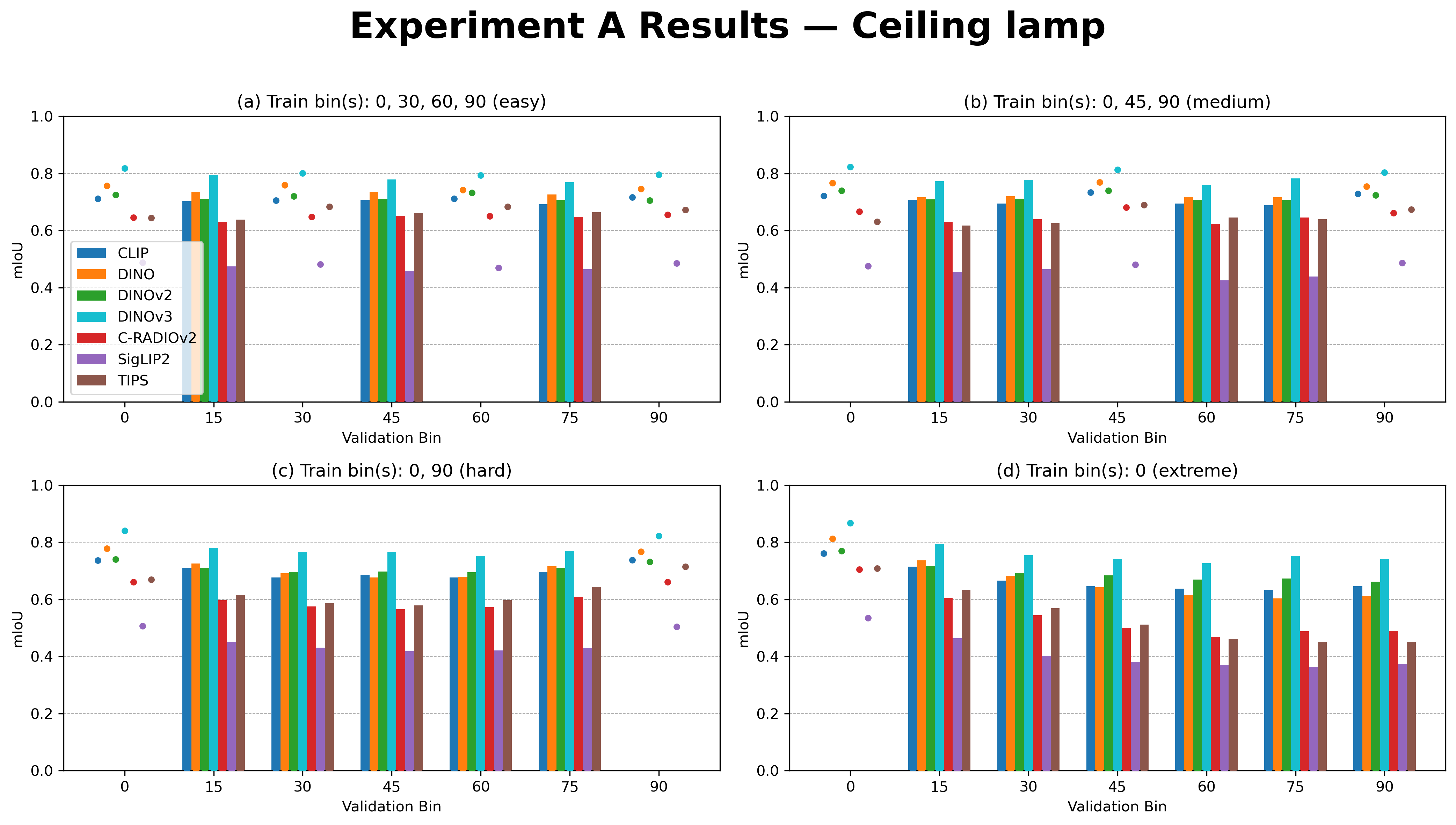}
        \caption{\textbf{Experiment A: ceiling lamp}. Performance trends closely mirror the general degradation observed in~\autoref{fig:expa_all_classes}, indicting the class has average viewpoint robustness. Degradation across bins follows the expected trend, with no unusual model failures.}
        \label{fig:expa_ceiling_lamp}
    \end{minipage}
    \hfill
    \begin{minipage}[t]{0.49\linewidth}
        \centering
        \includegraphics[width=\linewidth]{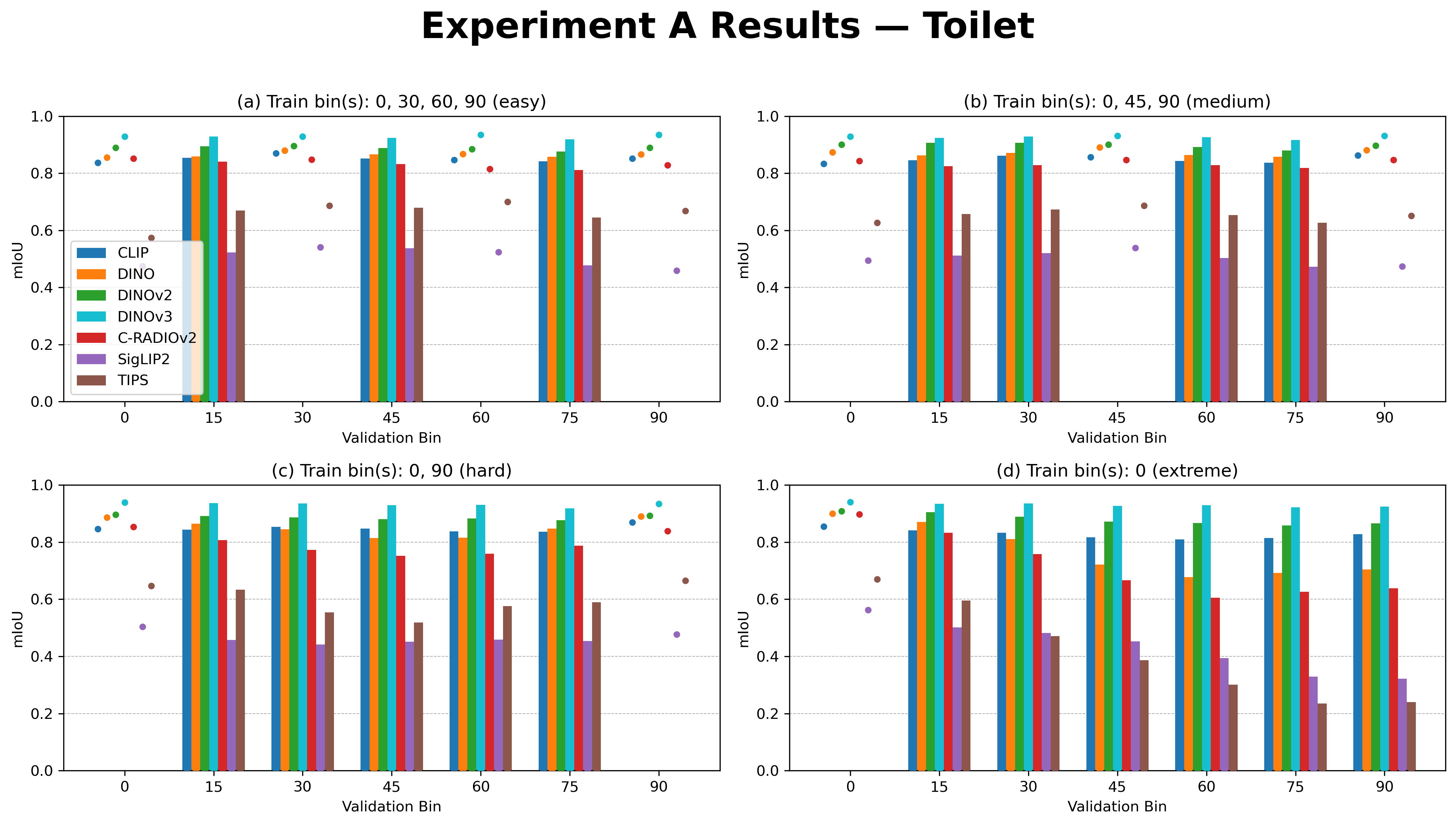}
        \caption{\textbf{Experiment A: toilet}. 
        The toilet class consistently performs above the overall trend in \autoref{fig:expa_all_classes}, with C-RADIOv2 showing the strongest boost, achieving increases of nearly 0.2 mIoU across difficulty levels.
        }
        \label{fig:expa_toilet}
    \end{minipage}
\end{figure}

\begin{figure}[H]
    \centering                                  
    \begin{minipage}[t]{0.49\linewidth}
        \centering
        \includegraphics[width=\linewidth]{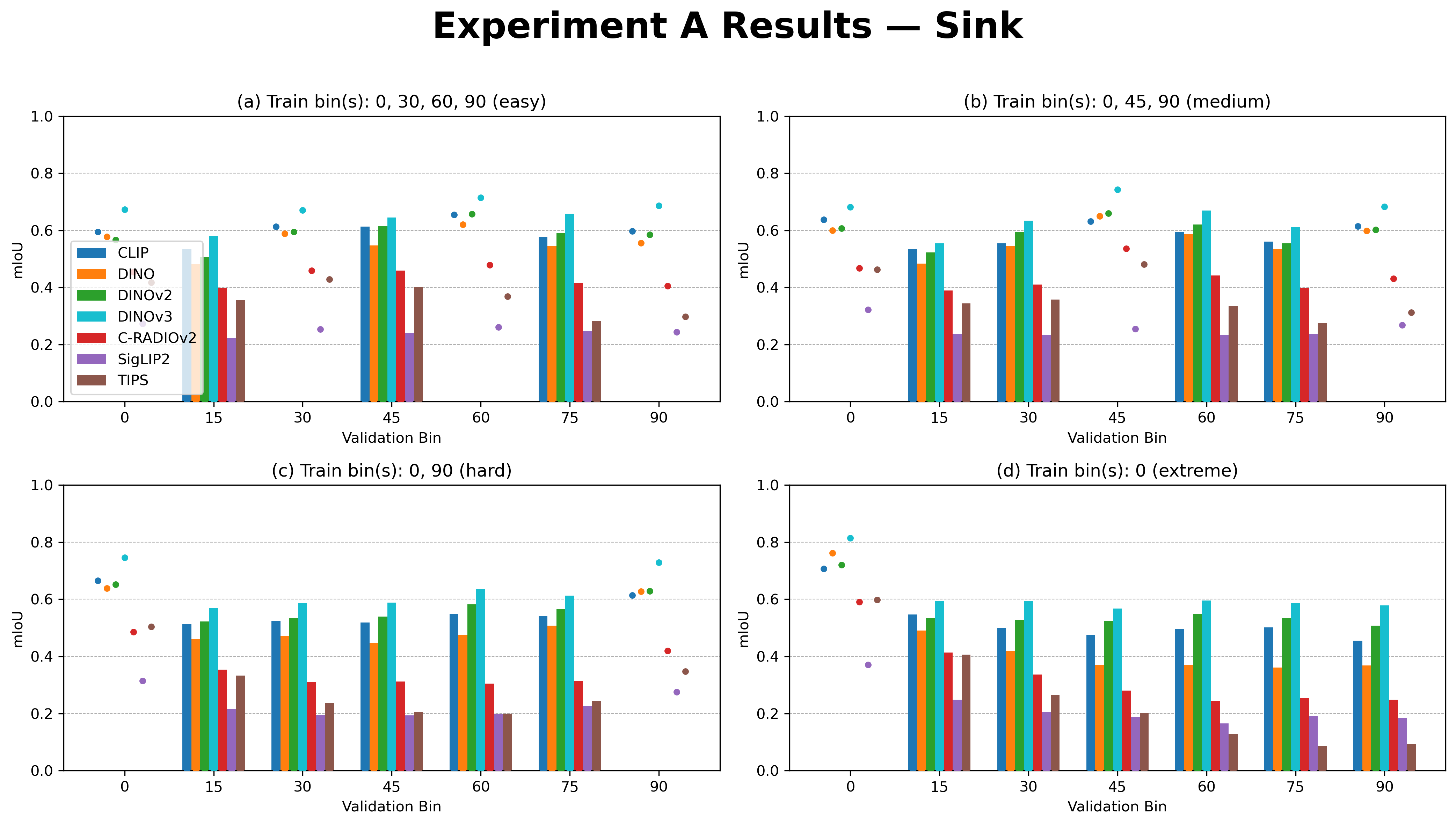}
        \caption{\textbf{Experiment A: sink}. 
        The sink class is one of the three worst-performing classes in the dataset. DINOv3 is the only model that performs well in this category. All other models show large negative gaps relative to \autoref{fig:expa_all_classes}, with SigLIP2 already underperforming in the easy setting by roughly 0.2 mIoU. Absolute scores generally remain low across all difficulty levels, and the degradation with viewpoint variation is steep. What geometric or appearance factors make sinks perform poorly is not immediately obvious, suggesting that sink segmentation is a broader failure mode worth investigating.
        }
        \label{fig:expa_sink}
    \end{minipage}
    \hfill
    \begin{minipage}[t]{0.49\linewidth}
        \centering
        \includegraphics[width=\linewidth]{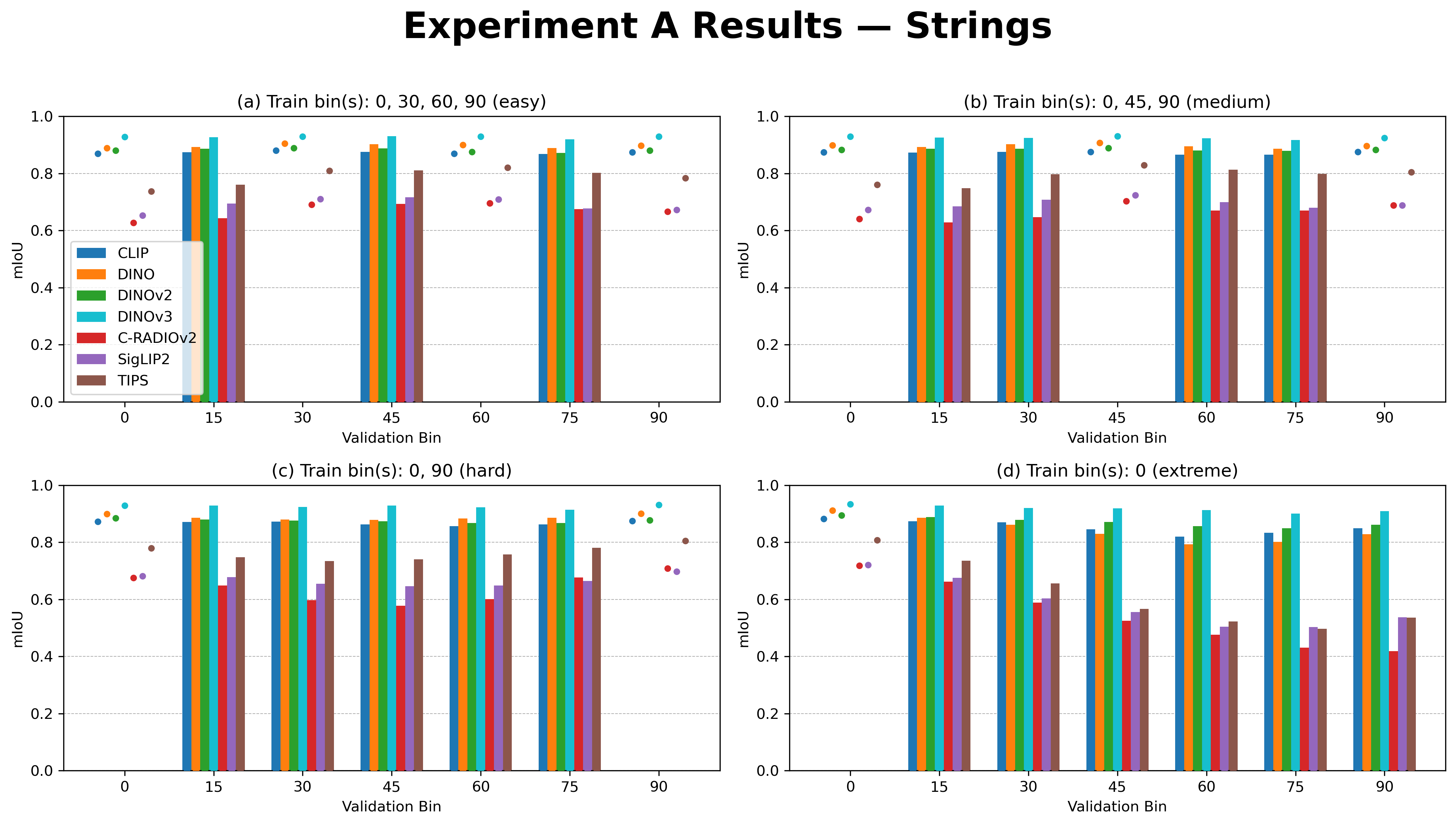}
            \caption{\textbf{Experiment A: strings}.
            The strings class generalizes reliably across all models, with performance declining smoothly as viewpoint difficulty increases, yet remaining consistently above the dataset average. Since this category spans multiple instruments (e.g. guitar, ukulele, guzheng), future work could investigate which visual attributes they share, which lead to strong viewpoint stability.
            }
        \label{fig:expa_strings}
    \end{minipage}
\end{figure}

\FloatBarrier
\newpage
\begin{figure}[H]
    \centering
    \begin{minipage}[t]{0.49\linewidth}
        \centering
        \includegraphics[width=\linewidth]{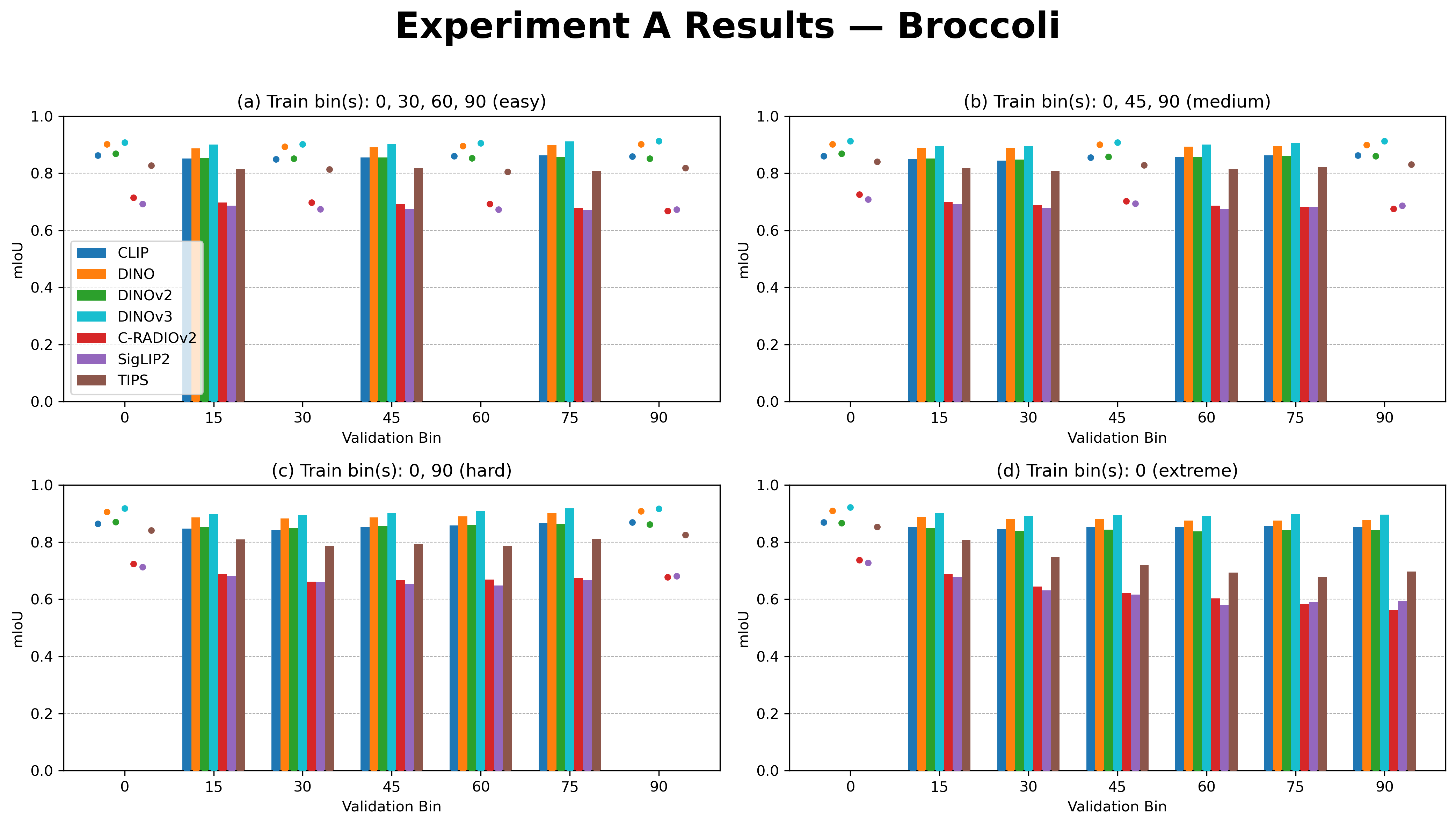}
        \caption{\textbf{Experiment A: broccoli}. 
        Models generalize well on the broccoli class, maintaining stable and above-average performance across unseen viewpoint bins. DINOv3 is consistently the top performer, the other DINO-based models and CLIP following closely. TIPS also performs strongly, trailing just behind the leaders in all but the Extreme setting and achieving notably higher scores than in \autoref{fig:expa_all_classes}. The results invite further investigation into which aspects of broccoli’s geometry support such robust viewpoint generalization.
        }
        \label{fig:expa_broccoli}
    \end{minipage}
    \hfill
    \begin{minipage}[t]{0.49\linewidth}
        \centering
        \includegraphics[width=\linewidth]{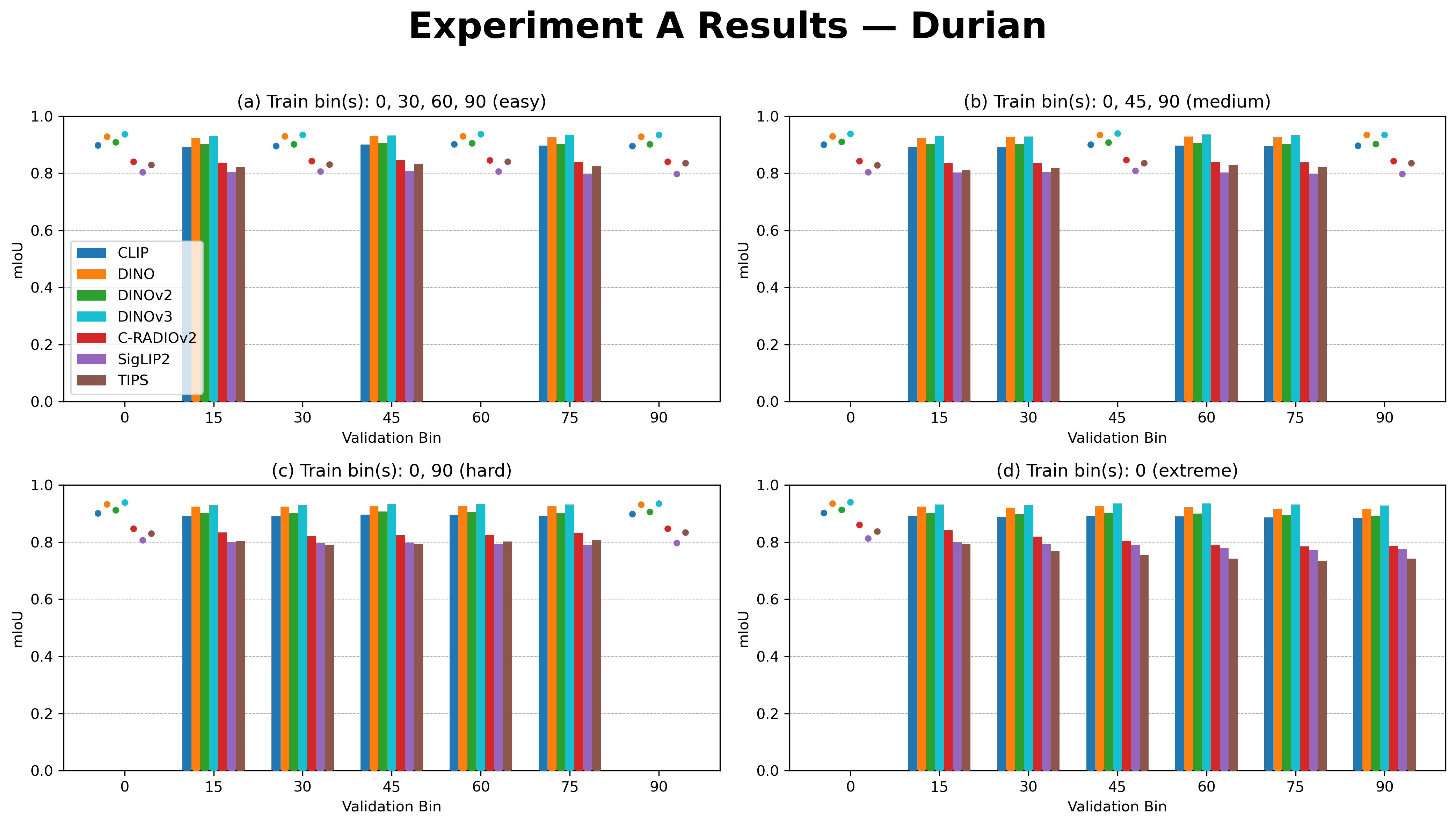}
        \caption{\textbf{Experiment A: durian}. 
        Durian is among the best-performing classes, showing high absolute mIoU (0.78--0.94) across most models and all difficulty levels. 
        This robustness is somewhat surprising: although the durian has a rich surface texture, its overall geometry is nearly spherical. It remains unclear whether the stable segmentation under viewpoint shifts stems from its distinctive visual signature or from contextual cues in the surrounding scene, such as the curtains highlighted in \autoref{fig:all_angles_objects}.
        }
        \label{fig:expa_durian}
    \end{minipage}
\end{figure}

\FloatBarrier
\newpage


\subsubsection{Bed category ground truth analysis}
\label{app:bed_gt_analysis}
We investigated the poor performance of the \textit{bed} class (as seen in \autoref{fig:12_bed}) and identified inconsistent or incomplete ground truth annotations as a likely cause.
\begin{figure}[H]
    \centering
    \begin{minipage}[t]{0.49\linewidth}
        \centering
        \includegraphics[width=\linewidth]{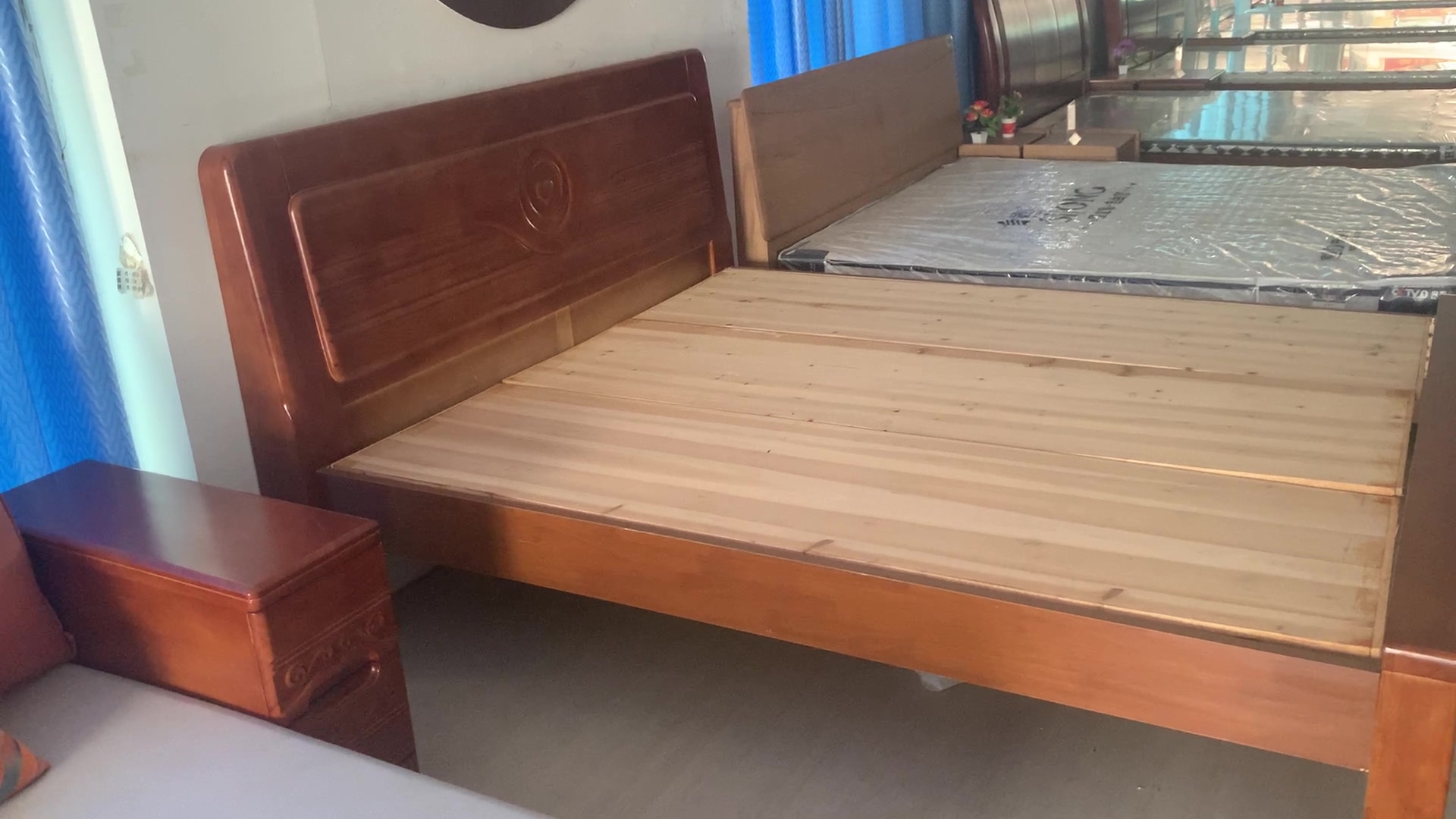}
        \caption{\textbf{Bed image example 1}. Original image used for annotation.}
        \label{fig:bed}
    \end{minipage}
    \hfill
    \begin{minipage}[t]{0.49\linewidth}
        \centering
        \includegraphics[width=\linewidth]{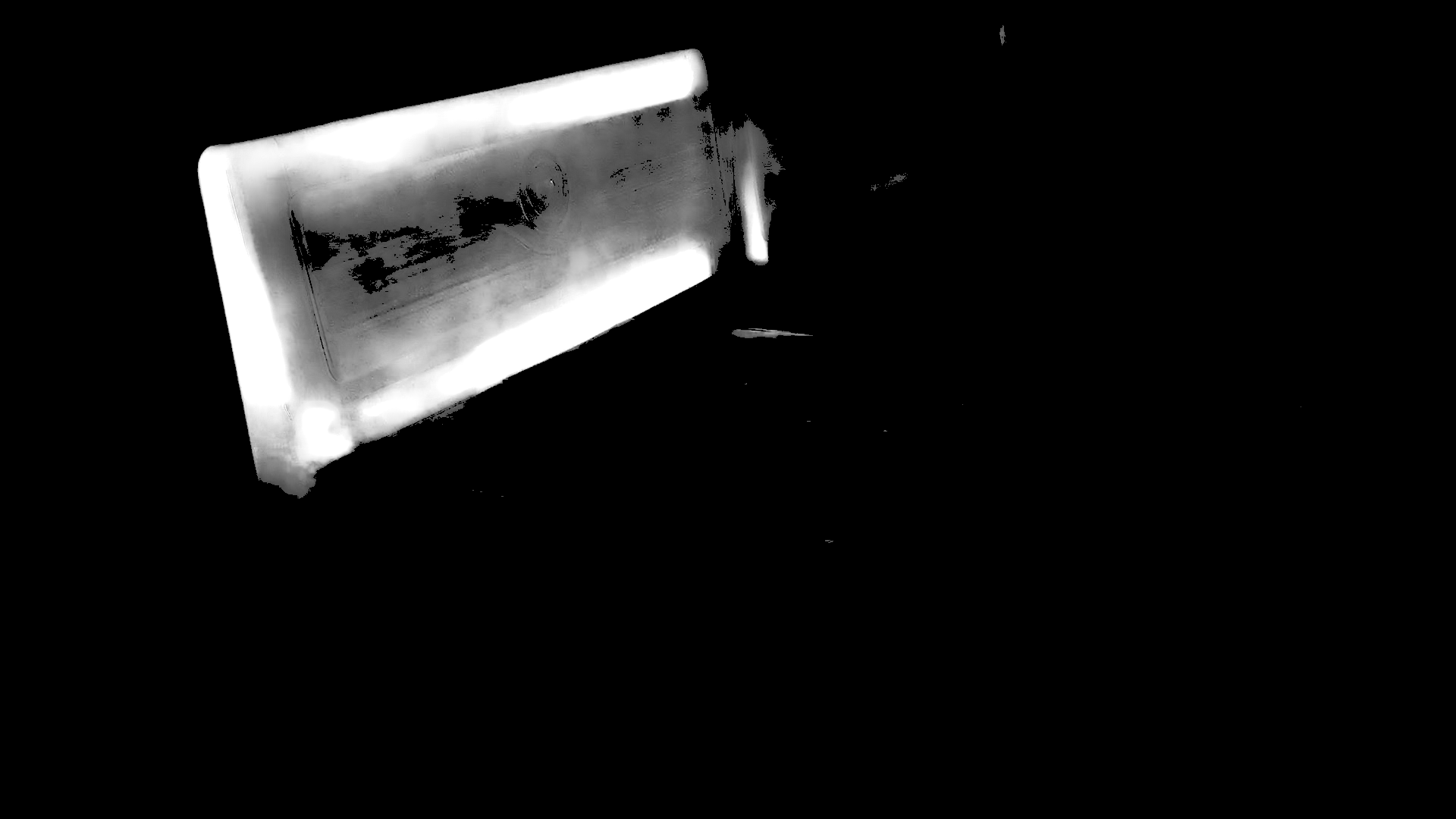}
        \caption{\textbf{Incomplete annotation}. Ground truth excludes parts of the bed (side railing, footboard) and ignores surrounding beds.}
        \label{fig:bed2}
    \end{minipage}
\end{figure}
    
\begin{figure}[H]
    \centering
    \begin{minipage}[t]{0.49\linewidth}
        \centering
        \includegraphics[width=\linewidth]{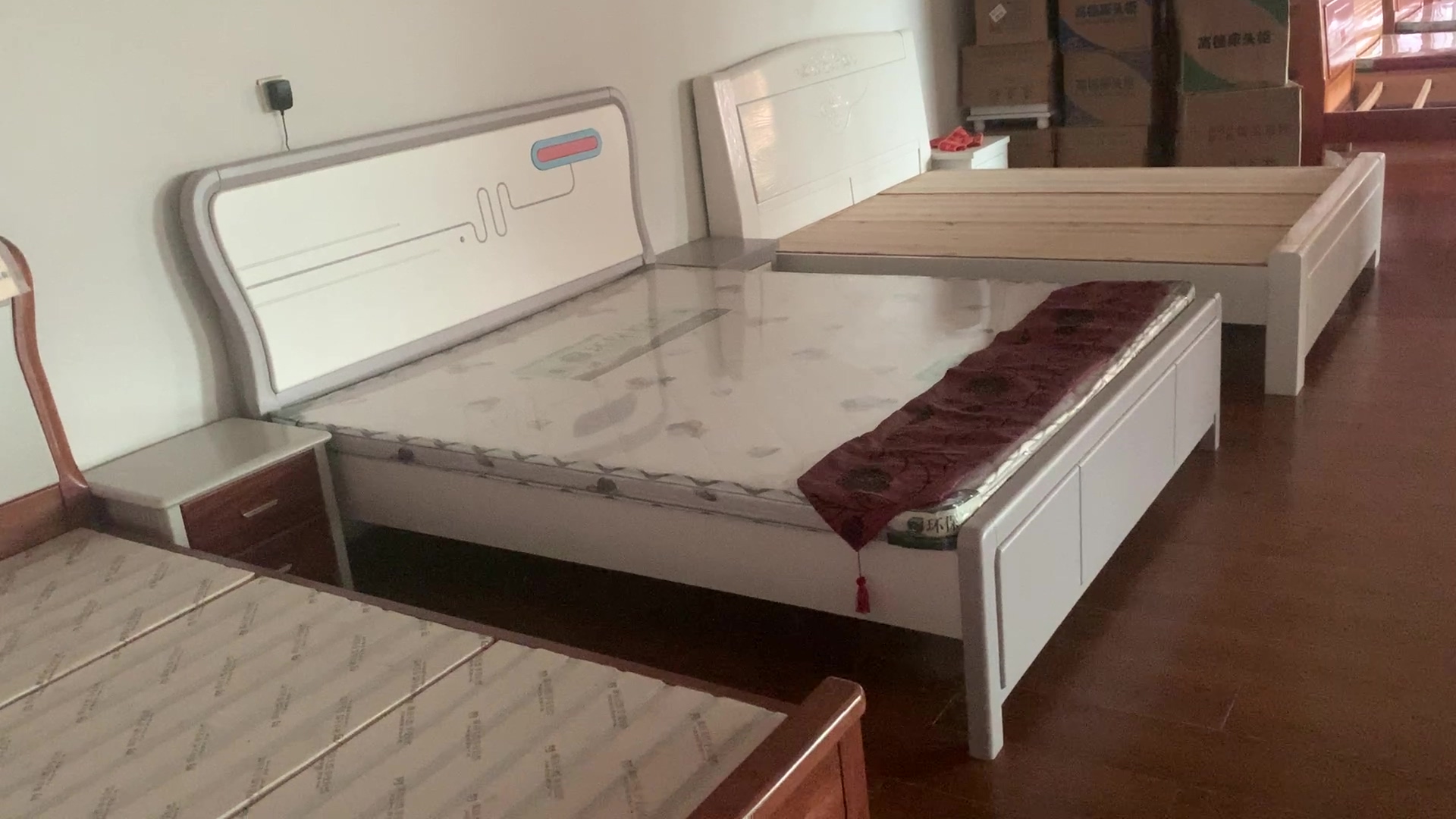}
        \caption{\textbf{Bed image example 2}. Original image used for annotation.}
        \label{fig:bed3}
    \end{minipage}
    \hfill
    \begin{minipage}[t]{0.49\linewidth}
        \centering
        \includegraphics[width=\linewidth]{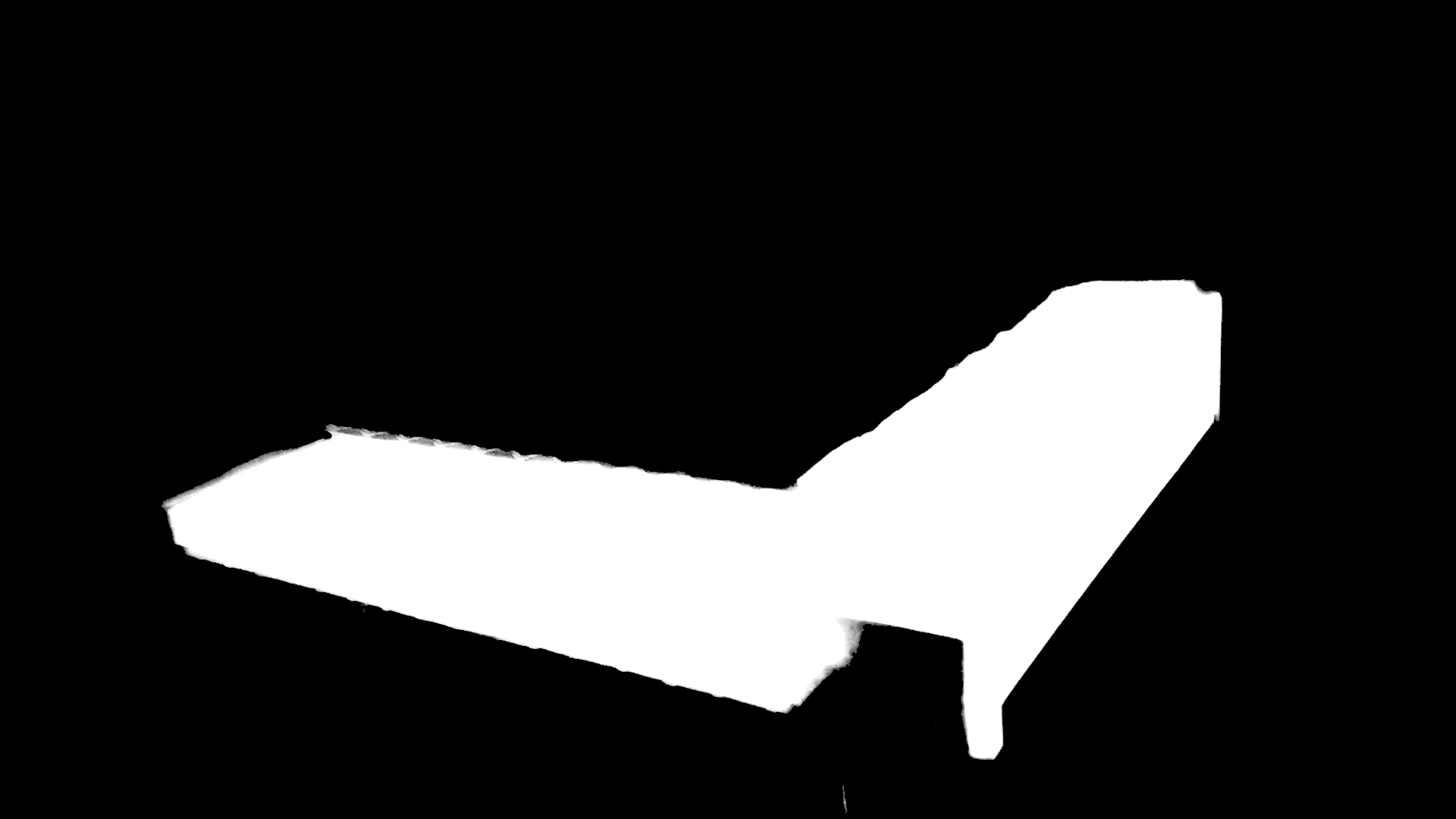}
        \caption{\textbf{Erroneous annotation}. The ground truth excludes large parts of the bed (headboard, mattress) and incorrectly includes the shadow between two beds.}
        \label{fig:bed4}
    \end{minipage}
\end{figure}
    
\begin{figure}[H]
    \centering
    \begin{minipage}[t]{0.49\linewidth}
        \centering
        \includegraphics[width=\linewidth]{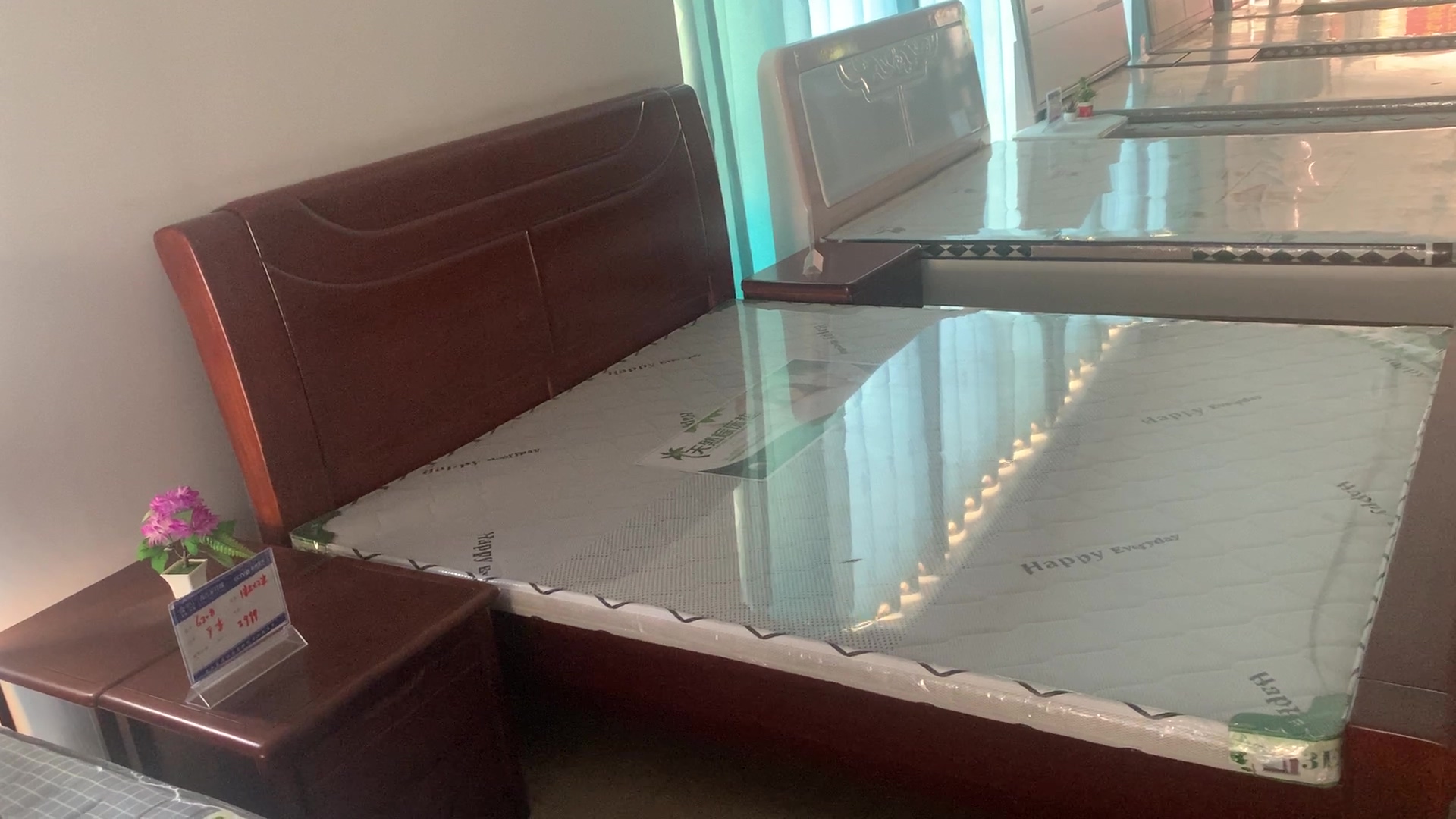}
        \caption{\textbf{Bed image example 3}. Original image used for annotation.}
        \label{fig:bed5}
    \end{minipage}
    \hfill
    \begin{minipage}[t]{0.49\linewidth}
        \centering
        \includegraphics[width=\linewidth]{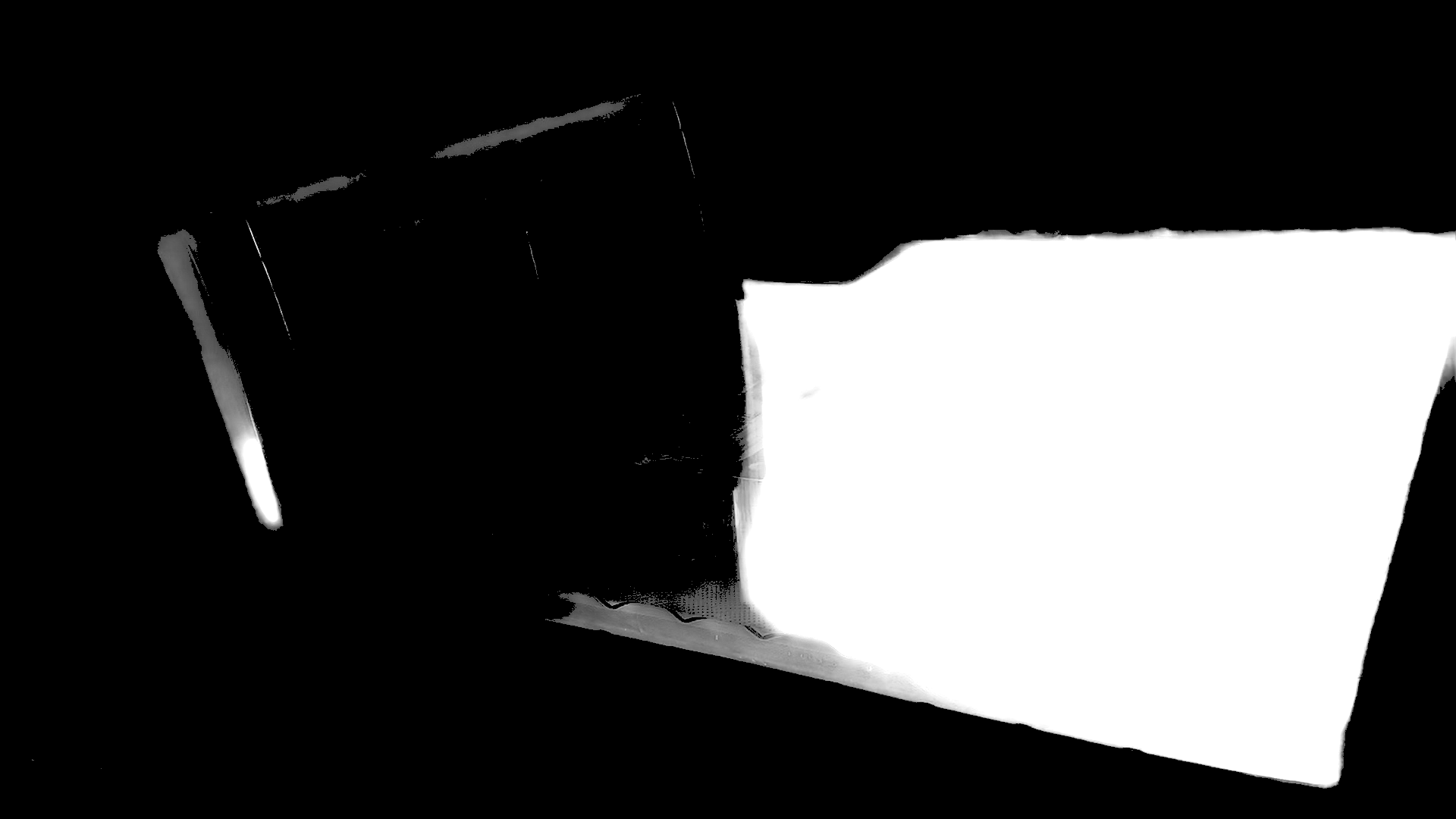}
        \caption{\textbf{Annotation error}. The ground truth omits the surrounding beds, incorrectly includes the gap between the main bed and the bed to its right, and only partially marks the headboard (side only).}

        \label{fig:bed6}
    \end{minipage}
\end{figure}

\FloatBarrier
\newpage

\subsection{Experiment B}
\label{app:expb}

\begin{figure}[H]
    \centering
    \includegraphics[width=0.95\linewidth]{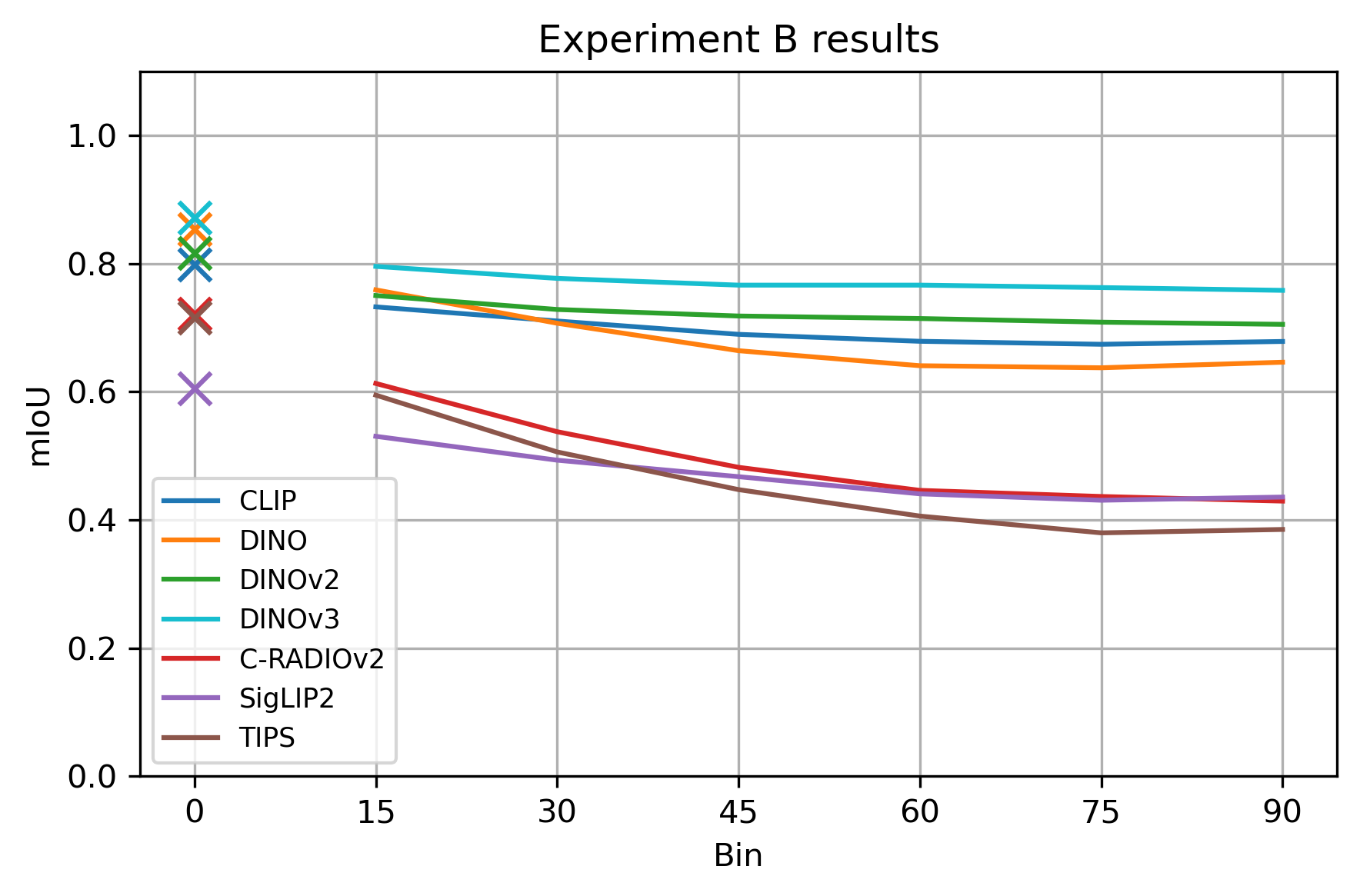}
    \caption{\textbf{Absolute mIoU under viewpoint shifts}.
Each curve shows the mIoU of a model evaluated on increasing validation bins, measured directly (without normalization to the 0\(^\circ\) baseline). The horizontal axis represents viewpoint angle relative to the training bin (0\(^\circ\)), and the vertical axis shows the segmentation performance via mIoU. All models show a gradual decrease as the angular distance grows, reflecting reduced correspondence between memory and query views. DINOv2 has the most consistent generalization across bins, maintaining a high mIoU across all bins. DINOv3 maintains the highest mIoU. By contrast, TIPS displays the steepest decline (dropping by more than 0.2 between 15 and 75), confirming its reduced robustness under large viewpoint changes.}
    \label{fig:expb_raw}
\end{figure}

As shown in~\autoref{fig:expb_norm} and~\autoref{fig:expb_raw}, most models exhibit a gradual performance decline as the distance of the validation bin from the training bin (0\(^\circ\)) increases. Visually, only C-RADIOv2 (ViT-B/16-CPE) and TIPS (ViT-B/14-HR)
exhibit sharp performance drops with increasing validation angles, indicating limited generalization capacity under extreme viewpoint shifts.

\FloatBarrier
\newpage

\subsection{Experiment C}
\label{app.ExpC}
In Experiment C, we evaluate the impact of memory bank size on model performance. \autoref{tab:hbird_memory_gain_combined} reports the absolute mIoU gains per difficulty level when increasing the memory bank size (a) from 320k to 640k, (b) from 640k to 1,024k, and (c) from 320k to 1,024k. 
This highlights how the sensitivity to additional memory varies across model architectures, while also showing diminishing returns as memory size increases.

To put these performance gains into practical context, we summarize the computational setup of Experiment~C. 
As mentioned in \autoref{App:Computational_Resources}, all experiments were executed on a single NVIDIA A100 GPU node with 4 GPUs. 
Across all configurations, runs took approximately 7--11 hours to complete, with larger memory banks consistently leading to longer runtimes. 
At larger memory sizes, runs encountered out-of-memory (OOM) errors unless additional measures were applied. 
We therefore fixed the batch size to 4 and enabled FAISS index sharding, reflecting the higher computational demands of larger memory configurations.

\begin{table}[H]
\centering
\footnotesize
\caption{\textbf{Performance gains from memory}. We report absolute mIoU improvements when increasing memory from (a) 320k to 640k, (b) 640k to 1{,}024k, and (c) 320k to 1{,}024k. Values are computed as differences from~\autoref{tab:memory_impact_performance}, with the final column showing the average gain across all 4 difficulty levels. 
}

\begin{tabular}{lccccc}
    \toprule
    & \multicolumn{5}{c}{\textbf{Gains}} \\
    \cmidrule(lr){2-6}
    \textbf{Model} & \textbf{Easy} & \textbf{Medium} & \textbf{Hard} & \textbf{Extreme} & \textbf{Average} \\
    \midrule
    \multicolumn{6}{c}{\textbf{(a) Memory: 320k $\rightarrow$ 640k}} \\
    \midrule
    CLIP ViT-B/16 & 0.017 & 0.016 & 0.017 & 0.013 & 0.016 \\
    DINO ViT-B/16 & 0.027 & 0.025 & 0.024 & 0.019 & 0.024 \\
    DINOv2 ViT-B/14 & 0.017 & 0.014 & 0.015 & 0.012 & 0.014 \\
    DINOv3 ViT-B/16 & 0.011 & 0.010 & 0.007 & 0.003 & 0.008 \\
    C-RADIOv2 ViT-B/16-CPE & 0.041 & 0.036 & \textbf{0.033} & \textbf{0.024} & \textbf{0.034} \\
    SigLIP2 B/16-512 & 0.035 & 0.030 & 0.029 & 0.023 & 0.029 \\
    TIPS ViT-B/14-HR & \textbf{0.044} & \textbf{0.038} & 0.032 & 0.016 & 0.033 \\
    \midrule
    \textbf{Average per task} & \textbf{0.027} & 0.024 & 0.022 & 0.016 & 0.022 \\

    \midrule
    \multicolumn{6}{c}{\textbf{(b) Memory: 640k $\rightarrow$ 1,024k}} \\
    \midrule
    
    CLIP ViT-B/16 & 0.010 & 0.008 & 0.007 & 0.007 & 0.008 \\
    DINO ViT-B/16 & 0.014 & 0.013 & 0.012 & 0.010 & 0.012 \\
    DINOv2 ViT-B/14 & 0.009 & 0.008 & 0.008 & 0.007 & 0.008 \\
    DINOv3 ViT-B/16 & 0.006 & 0.005 & 0.003 & 0.002 & 0.004 \\
    C-RADIOv2 ViT-B/16-CPE & \textbf{0.024} & \textbf{0.021} & \textbf{0.019} & \textbf{0.015} & \textbf{0.020} \\
    SigLIP2 B/16-512 & 0.023 & 0.020 & \textbf{0.019} & \textbf{0.015} & 0.019 \\
    TIPS ViT-B/14-HR & 0.023 & 0.019 & 0.017 & 0.009 & 0.017 \\
    \midrule
    \textbf{Average per task} & \textbf{0.015} & 0.014 & 0.012 & 0.009 & 0.013 \\

    \midrule
    \multicolumn{6}{c}{\textbf{(c) Memory: 320k $\rightarrow$ 1{,}024k}} \\
    \midrule
    
    CLIP ViT-B/16 & 0.026 & 0.024 & 0.024 & 0.020 & 0.024 \\
    DINO ViT-B/16 & 0.041 & 0.038 & 0.036 & 0.030 & 0.036 \\
    DINOv2 ViT-B/14 & 0.025 & 0.022 & 0.023 & 0.019 & 0.022 \\
    DINOv3 ViT-B/16 & 0.016 & 0.015 & 0.010 & 0.005 & 0.012 \\
    C-RADIOv2 ViT-B/16-CPE & 0.065 & \textbf{0.058} & \textbf{0.053} & \textbf{0.039} & \textbf{0.054} \\
    SigLIP2 B/16-512 & 0.058 & 0.050 & 0.047 & 0.038 & 0.048 \\
    TIPS ViT-B/14-HR & \textbf{0.067} & 0.057 & 0.049 & 0.025 & 0.049 \\
    \midrule
    \textbf{Average per task} & \textbf{0.043} & 0.038 & 0.035 & 0.025 & 0.035 \\

    \bottomrule
\end{tabular}
\label{tab:hbird_memory_gain_combined}
\end{table}

\FloatBarrier
\newpage



\section{Qualitative analysis}
\label{app:quali_analysis}

\begin{figure}[H]
    \centering
    \includegraphics[width=0.95\linewidth]{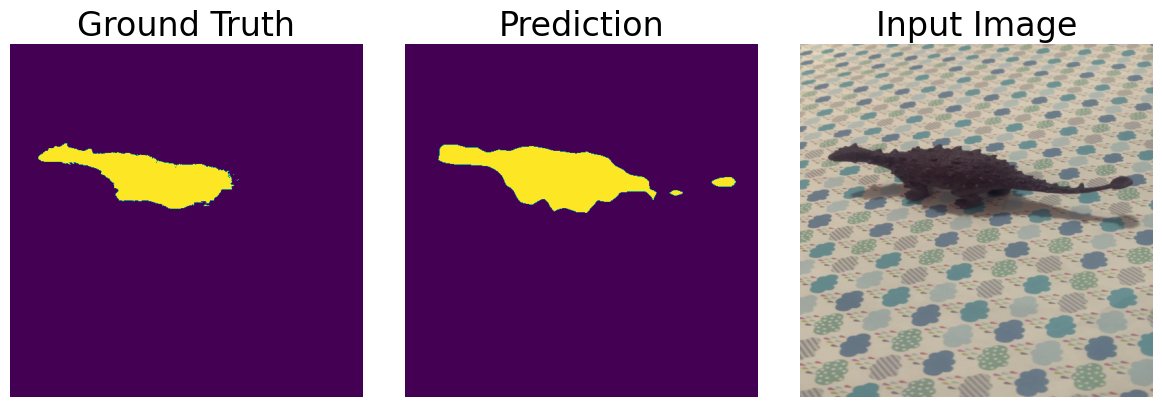}
    \caption{\textbf{Qualitative segmentation}. 
    Results shown using DINO. Left: input image. Center: predicted mask. Right: ground truth mask. 
    Interestingly, the prediction of the \textit{toy dragon} aligns more closely with visible object boundaries than the ground truth, which appears coarser and less accurate at fine details (e.g., tail, horns, stomach occlusion).}
    \label{fig:gt_pred_input_dino_triplet}

\end{figure}

\begin{figure}[H]

    \centering
    \includegraphics[width=0.95\linewidth]{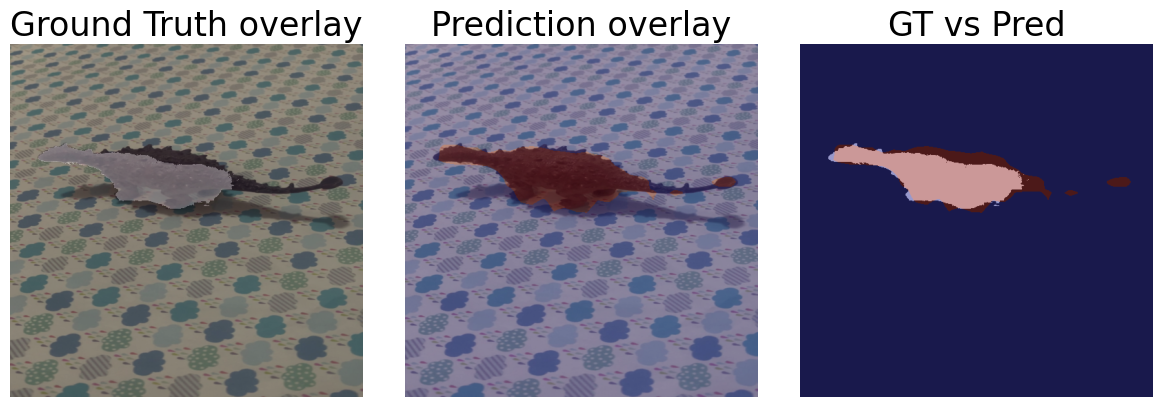}
    \caption{\textbf{Overlay comparison of ground truth and prediction}. 
    The same instance as in~\autoref{fig:gt_pred_input_dino_triplet} is shown for a better comparison. 
    Left: ground truth overlaid on the input image. 
    Center: prediction overlaid on the input image. 
    Right: difference visualization, with the ground truth overlaid on the predicted mask.}
    \label{fig:overlay_gt_pred_dino}

\end{figure}

\begin{figure}[H]
    
    \centering
    \includegraphics[width=0.35\linewidth]{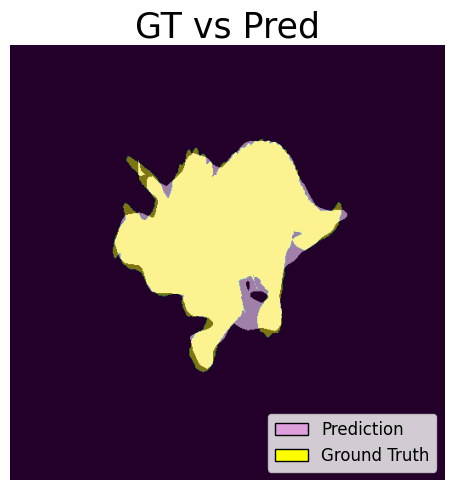}
    \caption{\textbf{Another overlay comparison of ground truth and prediction}. 
    Shown using DINO. 
    Yellow: ground truth mask. 
    Purple: predicted mask. 
    We note that discrepancies appear at fine-grained boundaries (e.g., tail, horns, nose) and in occluded or shadowed regions (e.g., under the stomach).}

    \label{fig:gt_vs_pred_triceratops}
\end{figure}



\end{document}